\documentclass[acmtog]{acmart}
\acmSubmissionID{203}

\usepackage{booktabs} 
\usepackage{colortbl}

\usepackage[ruled]{algorithm2e} 

\SetAlFnt{\small}
\SetAlCapFnt{\small}
\SetAlCapNameFnt{\small}
\SetAlCapHSkip{0pt}

\setcopyright{acmcopyright}\acmJournal{TOG}
\acmYear{2022}\acmVolume{41}\acmNumber{6}\acmArticle{195}\acmMonth{12}
\acmDOI{10.1145/3550454.3555447}

\def\framework{Text2Light}
\def\dataset{HDR360-UHD}

\definecolor{revised}{RGB}{0, 47, 167}

\definecolor{rred}{RGB}{245, 152, 153}
\definecolor{oorange}{RGB}{253, 205, 154}
\definecolor{yyellow}{RGB}{255, 255, 153}
\definecolor{green}{RGB}{169,209,142}
\definecolor{orange}{RGB}{239,182,143}

\begin{document}
\title{\framework: Zero-Shot Text-Driven HDR Panorama Generation}

\author{Zhaoxi Chen}
\email{zhaoxi001@ntu.edu.sg}
\orcid{0000-0003-3998-7044}
\affiliation{%
 \institution{S-Lab, Nanyang Technological University}
 \country{Singapore}}

\author{Guangcong Wang}
\email{guangcong.wang@ntu.edu.sg}
\orcid{0000-0002-6627-814X}
\affiliation{%
 \institution{S-Lab, Nanyang Technological University}
 \country{Singapore}}

\author{Ziwei Liu}
\authornote{corresponding author}
\email{ziwei.liu@ntu.edu.sg}
\orcid{0000-0002-4220-5958}
\affiliation{%
 \institution{S-Lab, Nanyang Technological University}
 \country{Singapore}}

\begin{abstract}
High-quality HDRIs (High Dynamic Range Images), typically HDR panoramas, are one of the most popular ways to create photorealistic lighting and 360-degree reflections of 3D scenes in graphics. Given the difficulty of capturing HDRIs, a versatile and controllable generative model is highly desired, where layman users can intuitively control the generation process. However, existing state-of-the-art methods still struggle to synthesize high-quality panoramas for complex scenes. In this work, we propose a zero-shot text-driven framework, \textbf{\framework}, to generate 4K+ resolution HDRIs without paired training data. Given a free-form text as the description of the scene, we synthesize the corresponding HDRI with two dedicated steps: \textbf{1)} text-driven panorama generation in low dynamic range (LDR) and low resolution (LR), and \textbf{2)} super-resolution inverse tone mapping to scale up the LDR panorama both in resolution and dynamic range. Specifically, to achieve zero-shot text-driven panorama generation, we first build dual codebooks as the discrete representation for diverse environmental textures. Then, driven by the pre-trained Contrastive Language-Image Pre-training (CLIP) model, a text-conditioned global sampler learns to sample holistic semantics from the global codebook according to the input text. Furthermore, a structure-aware local sampler learns to synthesize LDR panoramas patch-by-patch, guided by holistic semantics. To achieve super-resolution inverse tone mapping, we derive a continuous representation of 360-degree imaging from the LDR panorama as a set of structured latent codes anchored to the sphere. This continuous representation enables a versatile module to upscale the resolution and dynamic range simultaneously. Extensive experiments demonstrate the superior capability of \framework\ in generating high-quality HDR panoramas. In addition, we show the feasibility of our work in realistic rendering and immersive VR.
\end{abstract}

%
%

%
%


\begin{teaserfigure}
  \includegraphics[width=\textwidth]{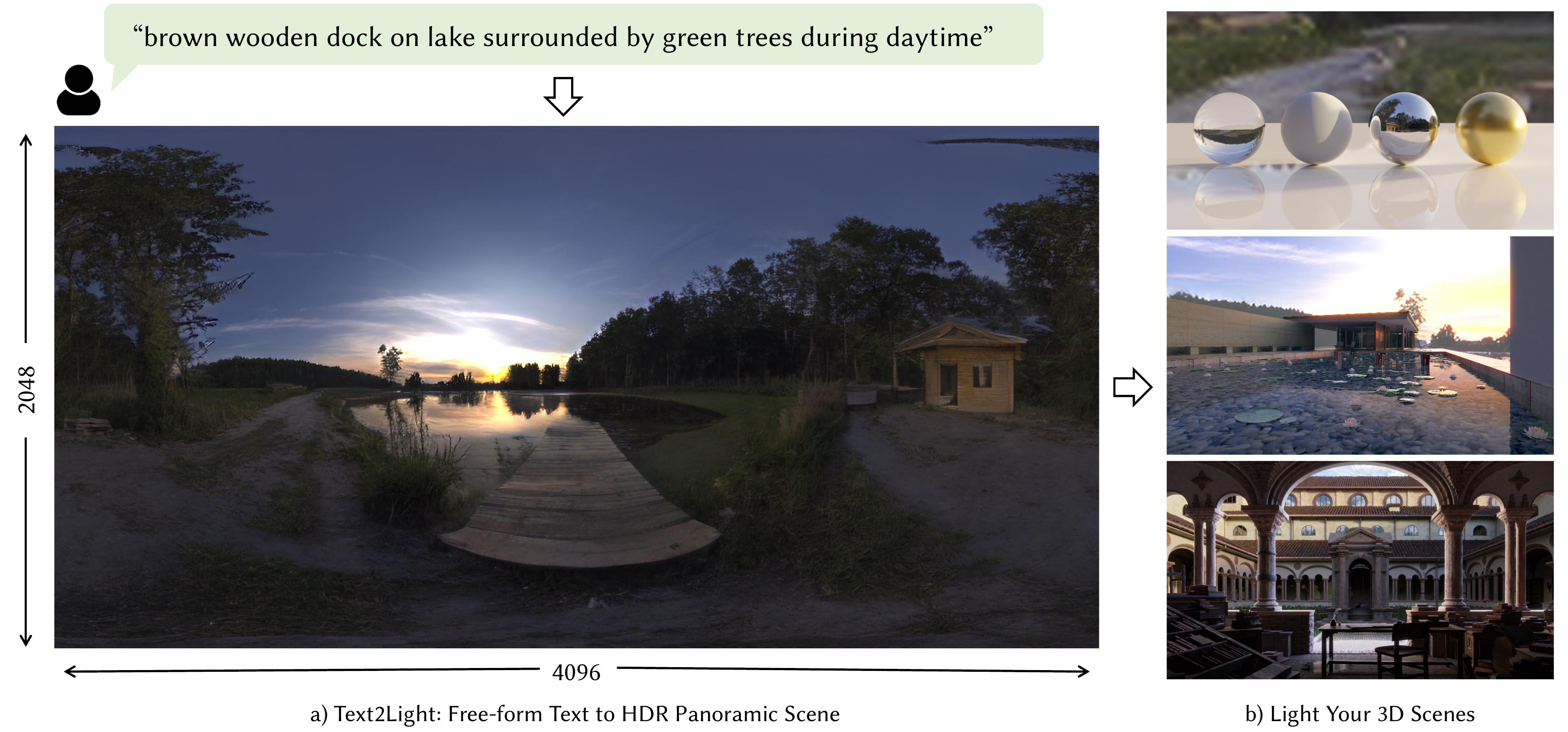}
  \caption{\textbf{Zero-shot text-driven HDR panorama generation}. a) \framework\ can generate the HDR panorama in 4K+ resolution using the free-form text solely. b) Our high-quality results can be directly applied to downstream tasks, \textit{e.g.,} light 3D scenes and immersive virtual reality.}
  \label{fig:teaser}
\end{teaserfigure}

\maketitle

\section{Introduction}
The emergence of the metaverse and virtual reality has fueled the demands for photorealistic rendering of 3D scenes~\cite{totalrelighting, chen2022relighting}. Among all techniques, HDRIs are one of the most efficient and popular ways to produce realistic illuminations of computer graphics (CG) scenes and immersed environment textures. A high-quality HDRI that could be directly used in the graphic pipeline is expected to have sufficient details with high resolution and high dynamic range. However, the acquisition of HDRIs is difficult, since the capture process is both expensive and constrained. It motivates us to design an automatic pipeline.

Unlike common scenery images, HDRIs represent the radiance of the scene in high dynamic range and 360-degree view, indicating that they are more diverse and fine-grained in content. Furthermore, from the perspective of application and interaction, it is also desired to provide intuitive control of the whole generation process for layman users. For instance, due to the massive amount of detailed information contained in a single HDRI, manually constructing and refining it by the user is tough work. A more intuitive way would be controlling the model with explicit textual descriptions, \textit{e.g.,} "brown wooden dock on lake surrounded by green trees during daytime", as shown in \textbf{Fig. \ref{fig:teaser}}, which is friendly for users without expert knowledge.

Despite the great potential, text-driven HDR panorama generation has been rarely explored in recent years due to the following challenges. 
\textbf{1) Resolution:} Existing state-of-the-art generative models~\cite{karras_style-based_2019, karras_analyzing_2020, karras_alias-free_nodate, lin_infinitygan_2021, esser_taming_2021} struggle to synthesize scene-level contents with ultra high resolution (4K+) and diverse yet sufficient details. 
\textbf{2) Coherence:} Unlike object-level images, scene-level panoramas often contain a large number of objects and structural layouts. It is hard to maintain the structural coherence and holistic semantics during panorama synthesis. 
\textbf{3) Aligning with texts:} It is expensive to collect panorama-text pairs for training. Generating scene-level content from free-form texts in a zero-shot manner remains unexplored. 
\textbf{4) HDR:} HDRIs cover a high dynamic range of lighting values, which often lead to unstable LDR-to-HDR regression. Coping with high dynamic range is still a non-trivial task during generation.

\begin{figure*}[t]
    \begin{center}
    \centerline{\includegraphics[width=2.01\columnwidth]{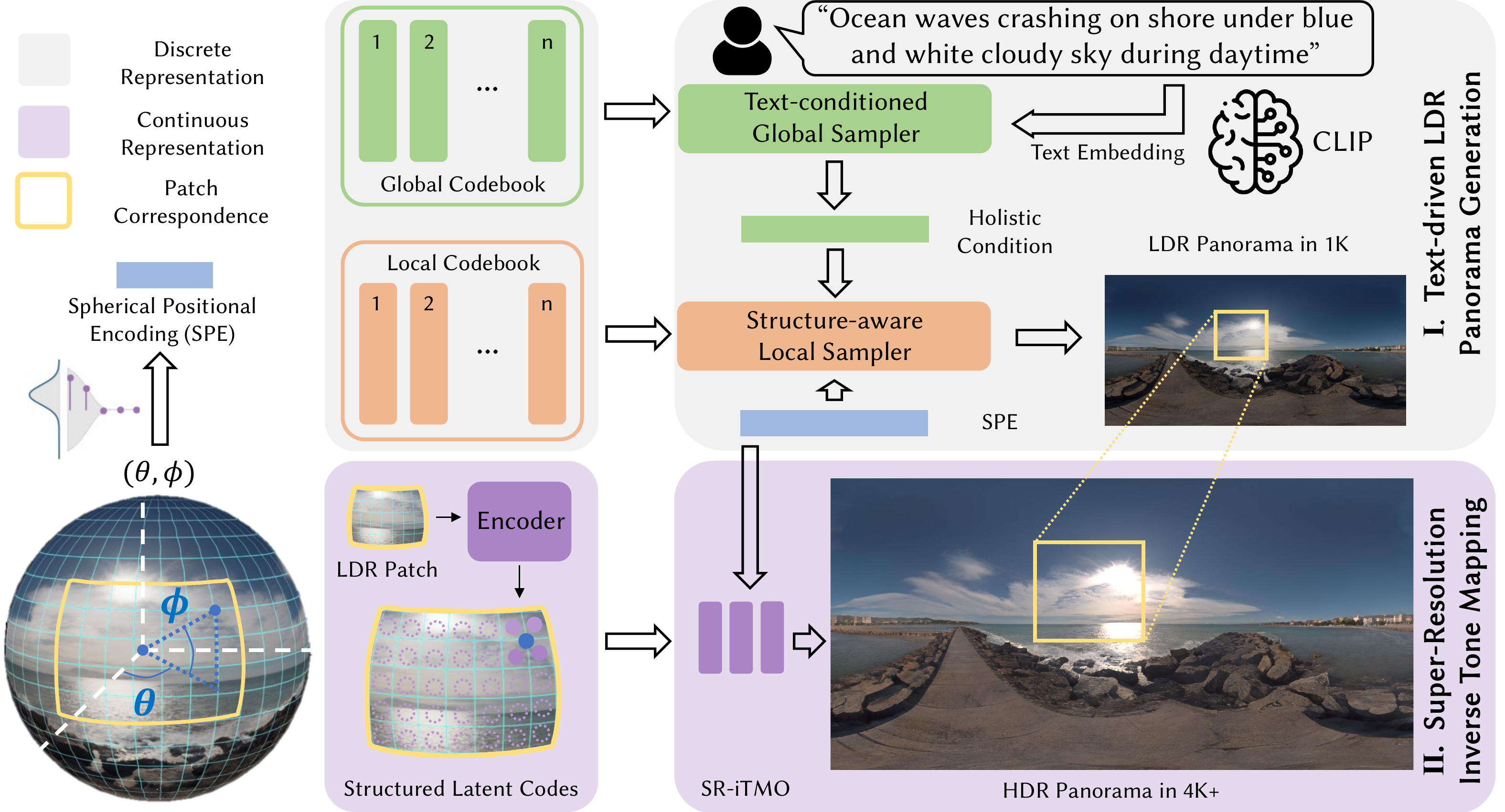}}
    \caption{\textbf{Overview of \framework}. We decompose the generation process of HDR panorama into two stages. \textbf{Stage I} translates the input text to LDR panorama based on a dual-codebook discrete representation. First, the input text is mapped to the text embedding by the pre-trained CLIP model~\cite{radford_learning_2021}. Second, a text-conditioned global sampler learns to sample holistic semantics from the global codebook according to the input text. Then, a structure-aware local sampler synthesizes local patches and composites them accordingly. \textbf{Stage II} upscales the LDR result from Stage I based on structured latent codes as continuous representations. We propose a novel Super-Resolution Inverse Tone Mapping Operator (SR-iTMO) to simultaneously increase the spatial resolution and dynamic range of the panorama.
    }\label{fig:overview}
    \end{center}
\end{figure*}

To solve these problems, we propose, \textbf{\framework}, the first framework for zero-shot text-driven HDR panorama generation, as shown in \textbf{Fig. \ref{fig:overview}}. Due to the complexity of HDR panorama in detailed textures, spherical structure, and high dynamic range, we decompose the generation process into two dedicated stages. \textbf{1) Stage I} generates a 360-degree panorama in low dynamic range (LDR) and low resolution (LR) given a description of the scene as the input. \textbf{2) Stage II} upscales the result of Stage I both in resolution and dynamic range by a super-resolution inverse tone mapping operator (SR-iTMO). \textbf{Stage I} focuses on preserving the structural coherence and holistic semantics without using text-panorama paired training data, and \textbf{Stage II} aims at generating high-quality HDRIs with ultra high resolution (4K+) and high dynamic range. With the proposed two-stage framework, users can generate a high-quality HDR panorama in 4K+ resolution using only the text description. The results can be used in the rendering engine to produce realistic illuminations and immersed environment textures, as shown in \textbf{Fig. \ref{fig:teaser}}.

In Stage I, given the extremely high diversity of environmental textures, we leverage the codebook, which is commonly used in vector-quantized models~\cite{oord_neural_2018, esser_taming_2021}, as a discrete representation of LDR textures. Specifically, to handle both the global coherence and local details, we introduce a new dual-codebook representation in a coarse-to-fine manner for distinct objectives. The global codebook aims to represent the coarse global appearance and scene-level semantics, while the local codebook focuses on encoding fine-grained details. 
With this hierarchical discrete representation, the task for Stage I is formulated as sampling appropriate embeddings from dual codebooks given the input text. 
To this end, we first derive the text embedding from the scene description with the aid of the pre-trained CLIP model~\cite{radford_learning_2021}. 
Then, a text conditioned global sampler, which learns to align descriptions with scenes without text-panorama paired annotations, is employed to sample the holistic condition from the global codebook. 
Furthermore, a structure-aware local sampler is leveraged to sample features from the local codebook subject to the holistic condition and spherical positional encoding, which are finally decoded to LDR panoramas.

In Stage II, we treat each panorama as a continuous field distributed on a sphere, which serves as a complementary representation to codebooks in Stage I. We first encode the LDR panorama in a patch-wise manner to structured latent codes that are anchored on the sphere. Then, for any position $(\theta, \phi)$ on the sphere, we derive its latent feature via interpolating among its nearest neighbors, which is further fed into an MLP-based SR-iTMO to produce the corresponding HDR value in higher resolution.

To facilitate scene generation, we construct a high-quality dataset, \dataset, which contains 4393 HDR panoramas of both outdoor and indoor scenes, ranging from 4K to 8K resolution. 

We summarize our contributions as follows: 

\begin{itemize}
    \item We build a zero-shot text-driven framework, \framework, for HDR panorama generation, which produces photo-realistic illuminations and scene textures in 3D CG scenes controlled by free-form texts.
    \item We propose a dual-codebook discrete representation to describe both global scene-level semantics and fine-grained details. With dual codebooks, we introduce a text-conditioned global sampler, a structure-aware local sampler, and a spherical positional encoder to synthesize high-quality panoramas aligned with texts.
    \item We propose an efficient super-resolution inversed tone mapping operator to upscale the resolution and the dynamic range of images jointly.
    \item \framework\ shows the unprecedented ability to generate HDR panoramas in ultra 4K resolution with high fidelity.
\end{itemize}

\section{Related Work}

\paragraph{High-resolution Generative Models.}
Recent years have witnessed the rapid growth of generative models in high-resolution image synthesis~\cite{karras_style-based_2019, karras_analyzing_2020, karras_alias-free_nodate, razavi2019generating, gur2020hierarchical, skorokhodov_aligning_2021, chang_maskgit_2022, ho_denoising_2020, rombach_high-resolution_2021, bond-taylor_unleashing_2021, huang_multimodal_2021, wang2022stylelight}. StyleGAN~\cite{karras_style-based_2019} and its variants (e.g., StyleGAN2 and StyleGAN3)~\cite{karras_analyzing_2020, karras_alias-free_nodate} have substantial improvements in resolution and level of details, succeeding in synthesizing photorealistic images with a limited field of view. Yet, such methods are not able to generate rich and diverse content at the scene level. Besides, these methods train on the full image frame,  which makes them memory intensive when scaling up. A further limitation is that they fail to keep the fidelity of both holistic appearance and fine-grained details due to the limited receptive field. In contrast, patch-based methods~\cite{lin_coco-gan_2020, lin_infinitygan_2021, esser_taming_2021, skorokhodov_aligning_2021} are trained on small image patches to get rid of computational resource bottlenecks, succeeding in generating high-resolution scenery images. However, Taming Transformer~\cite{esser_taming_2021} can only synthesize megapixel images with dense conditions, \textit{e.g.,} segmentation maps or depth maps. InfinityGAN~\cite{lin_infinitygan_2021} claims to be able to generate infinite pixels, yet it fails to maintain the holistic semantic of the scene, which leads to low fidelity when zooming in. 

\paragraph{Text to Image.}
Most existing text-to-image models~\cite{ding_cogview_2021, nichol_glide_2022, dalle, ramesh_hierarchical_nodate, rombach2021ldm} are learned with supervision, using text-image pairs. Recently, the pre-trained CLIP model~\cite{radford_learning_2021} enables zero-shot text-to-image generation and manipulation~\cite{zhou_lafite_2022,kim_diffusionclip_2022, patashnik_styleclip_2021, schaldenbrand_styleclipdraw_2021, wang_clip-gen_2022, liu_fusedream_2021, hong2022avatarclip, jiang2022text2human, rdm} without paired data. However, directly guiding the generative model using CLIP may suffer from the semantic gap between the image and text embedding spaces in complex scenes. Given the complex semantics of a single scene, aligning the actual semantics of the input scene description with the generated images is still a non-trivial task.

\paragraph{Inverse Tone Mapping.}
In the context of photorealistic rendering, it is crucial for the environmental texture to have a high dynamic range that parameterizes the radiance of the real world. Intuitively, one could reconstruct HDR assets from the LDR counterpart via inverse tone mapping~\cite{banterle_inverse_2006}. In recent years, many convolutional methods~\cite{lee_deep_2018, wang_deep_2021, marnerides_expandnet_2019, yu_luminance_2021, zhang_learning_2017, liu_single-image_2020, chen_hdr_2021, raipurkar_hdr-cgan_2021, kim2020jsi, wei_beyond_2021} aim to generate high-quality HDR from LDR images. 
However, these works leverage fully convolutional network which is not robust to different scales~\cite{xu2014scale}. We propose to represent HDR panorama as a continuous field on a sphere. With this novel representation, we could efficiently perform inverse tone mapping from arbitrary resolution to arbitrary resolution simply by multi-layer perceptrons. 

\section{Text2Light}
Our goal is to generate a high-quality HDR panorama in 4K+ resolution given a scene description as the input. The difficulty of this task has three folds, i.e., scene-level coherent textures with high-resolution, zero-shot text-driven generation, and HDR mapping. Existing state-of-the-art generators, such as StyleGAN3~\cite{karras_alias-free_nodate} and InfinityGAN~\cite{lin_infinitygan_2021}, still struggle to generate realistic scene-level panoramas in 1K resolution. We present \textbf{\framework}, a novel controllable framework to generate a high-quality HDR panorama from a free-form text as the scene description, which is illustrated in \textbf{Fig. \ref{fig:overview}}. 

\framework\ mainly consists of two stages. In \textbf{Stage I} (Sec. \ref{sec:stage1}), we propose a hierarchical framework to generate LDR panorama driven by the text $T$, which is based on a dual-codebook discrete representation. 
Driven by the CLIP~\cite{radford_learning_2021} model, a text-conditioned global sampler learns to sample holistic semantics from the global codebook conditioned on the input text. Furthermore, a structure-aware local sampler learns to synthesize LDR panoramas patch-by-patch, guided by the holistic condition.
In \textbf{Stage II} (Sec. \ref{sec:stage2}), we propose a new super-resolution inverse tone mapping operator, termed as SR-iTMO. It upscales the result from Stage I both in spatial resolution and dynamic range to HDR 4K+ version using a continuous representation. With the SR-iTMO, \framework\ generates ultra-high-resolution and high-dynamic panoramas, which can be directly used in rendering engines or VR applications. Besides, different from common scenery images with limited field of views, panoramas can be regarded as 360-degree spherical fields. To integrate this prior into panorama generation, we introduce a spherical positional encoding (Sec. \ref{sec:spe}), or SPE for short, to represent 360-degree scenes in the entire pipeline.

\subsection{Spherical Positional Encoding}
\label{sec:spe}
Considering the intrinsic spherical structural property of panoramic images, we need a careful design to incorporate such a prior into panorama synthesis, which allows the framework to learn position-sensitive signals of a 3D scene. Given a panoramic image $I \in \mathbb{R}^{H\times W\times 3}$, we represent each pixel as a 3D point on an unit sphere $S$. Let $I(i,j)$ be the pixel value at $(i,j)$ in $I$ where $i \in \{1, ..., H\}$ and $j \in \{1, ..., W\}$, we represent its spherical counterpart as $S(\theta, \phi, r)$:
\begin{equation}
\label{eq:spherical}
\begin{aligned}
    &S(\theta, \phi, r) = I(i, j), \\
    &\theta = (2i/H - 1)\pi, \\
    &\phi = (2j/W - 1)\pi/2,\\
    &r=1.
\end{aligned}
\end{equation}

Intuitively, Eqn. \ref{eq:spherical} inverses the process of the equirectangular projection which converts a 360-degree raw image into a 2D panorama. The center of the sphere $S$ is exactly the position of the camera that takes the shot in the 3D world. 

Moreover, as deep networks are biased towards learning lower frequency functions~\cite{rahaman2019spectral}, we leverage Fourier positional encoding $\gamma(\cdot)$ to map $(\theta, \phi)$ to a higher dimensional space:
\begin{equation}
\label{eq:spe}
\begin{aligned}
    \gamma(\theta) &= [\sin(2^0\pi\theta), \cos(2^0\pi\theta),..., \sin(2^{L-1}\pi\theta), \cos(2^{L-1}\pi\theta)], \\
    \gamma(\phi) &= [\sin(2^0\pi\phi), \cos(2^0\pi\phi),..., \sin(2^{L-1}\pi\phi), \cos(2^{L-1}\pi\phi)],
\end{aligned}
\end{equation}
where $L$ is set to 4 in our experiments.

Such a representation acts as a strong inductive bias in our model, in the way of preserving high-frequency details and spherical structure in Stage I and introducing geographical priors in Stage II. We will introduce the usage of this representation in related sections.

\subsection{Stage I: Text-driven LDR Panorama Generation}
\label{sec:stage1}

\begin{figure*}[t]
    \begin{center}
    \centerline{\includegraphics[width=1.98\columnwidth]{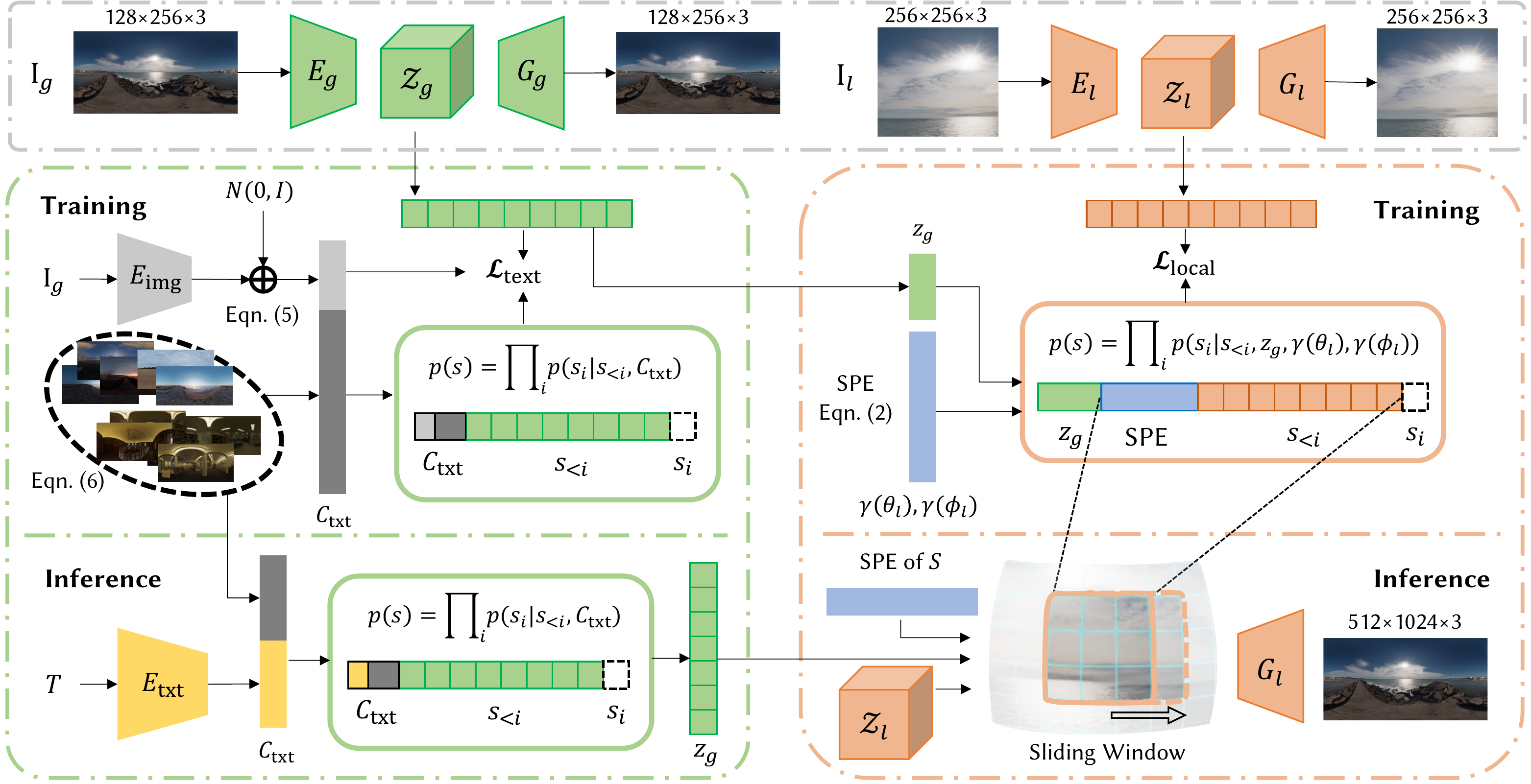}}
    \caption{\textbf{Overview of Stage I}. In Stage I, we aim to generate an LDR panorama using solely natural language descriptions. The hierarchical framework can be decomposed into three components. \textbf{\textcolor{gray}{i) Dual-codebook Discrete Representation:}} Building dual codebooks, $\mathcal{Z}_g, \mathcal{Z}_l$, to represent the holistic appearance and local fine-grained details. \textbf{\textcolor{green}{ii) Text-conditioned Global Sampler:}} Sampling holistic semantics from the global codebook according to the input text. \textbf{\textcolor{orange}{iii) Structure-aware Local Sampler:}} Synthesizing LDR panorama guided by the holistic condition in a structure-aware manner. The process of training and inference is introduced, respectively.
    }\label{fig:stage1}
    \end{center}
\end{figure*}

In Stage I, we propose a hierarchical framework that focuses on generating a low resolution ($1024\times 512$) panorama in low dynamic range using solely natural language descriptions. 
It can be decomposed into three components. \textbf{i) Dual-codebook representation:} Building dual codebooks to represent the holistic appearance and local fine-grained details. \textbf{ii) Text-conditioned global sampler:} Sampling holistic semantics from the global codebook conditioned on the input text. \textbf{iii) Structure-aware local sampler:} Synthesizing structure-aware LDR panoramas guided by the holistic condition patch-by-patch. An overview of Stage I is shown in \textbf{Fig. \ref{fig:stage1}}. 

\subsubsection{Image Tokenization}
Vector-Quantized Variational AutoEncoder (VQVAE)~\cite{oord_neural_2018} is proposed to learn a discrete representation, a codebook $\mathcal{Z}$, by reconstructing images. Given an input image $I$, VQVAE leverages an encoder ${E}$ to obtain the continuous feature $\hat{z}$, \textit{i.e.}, $\hat{z} = {E}(I) \in \mathbb{R}^{h\times w\times c_z}$. Then a quantization is performed onto its closest codebook entry $z_k$ to obtain the discrete representation $z_q$:
\begin{equation}
\label{eq:quant}
z_q = \underset{z_k \in \mathcal{Z}}{\mathrm{argmin}}||\hat{z} - z_k||_2.
\end{equation}
Then the image is reconstructed by a decoder ${G}$ as $\hat{I} = {G}(z_q)$. To improve the perceptual quality at a high compression rate~\cite{esser_taming_2021}, a discriminator ${D}$~\cite{isola2017image} is also introduced to discriminate between real and generated image patches. Thus, the overall training objective is:
\begin{equation}
\label{eq:vqgan}
\mathcal{L} = \mathcal{L}_\mathrm{rec} + ||\mathrm{sg}(z_q) - \hat{z}||^2_2 + ||\mathrm{sg}(\hat{z}) - z_q||^2_2 + \mathcal{L}_\mathrm{GAN},
\end{equation}
where $\mathcal{L}_\mathrm{rec}$ is the perceptual loss~\cite{zhang2018unreasonable} between $I$ and $\hat{I}$, $\mathcal{L}_\mathrm{GAN} = \log {D}(I) + \log (1-{D}(\hat{I}))$, and $\mathrm{sg}(\cdot)$ denotes the stop-gradient operation.

\subsubsection{Dual-codebook Discrete Representation}\label{sec:discrete}
Given that textures of outdoor and indoor scenes are diverse and complicated, a single VQVAE (or StyleGAN) often fails to synthesize coherent appearance and sufficient local details simultaneously (see \textbf{Fig. \ref{fig:ldr-gen}}). 
Therefore, we derive a hierarchical discrete representation of LDR panoramas as dual codebooks. Specifically, to represent holistic semantics and coarse but global features, we first build a \textit{global codebook} $\mathcal{Z}_g$ via training a VQVAE (Eqn. \ref{eq:vqgan}) to reconstruct a downsampled panorama $I_g \in \mathbb{R}^{128\times 256\times 3}$. Let $E_g$ and $G_g$ be the corresponding encoder and decoder, respectively. The encoder ${E}_g$ turns $I_g$ into a continuous feature $\hat{z}_g = {E}_g(I_g) \in \mathbb{R}^{8\times16\times c_z}$, then the coarse feature $z_g$ is derived from $\mathcal{Z}_g$ via quantization (Eqn. \ref{eq:quant}). Furthermore, to represent sufficient details and diverse textures across different scenes, we build a \textit{local codebook} $\mathcal{Z}_l$ via training another VQVAE to reconstruct high-resolution patches $I_l \in \mathbb{R}^{256\times 256\times 3}$. Let $E_l$ and $G_l$ be the corresponding encoder and decoder, we can obtain a fine-grained feature $z_l \in \mathbb{R}^{16\times 16\times c_z}$ by $E_l$ and $\mathcal{Z}_l$. During training, the patch $I_l$ is randomly cropped from the full LDR panorama $I \in \mathbb{R}^{512\times 1024\times 3}$, and $c_z$ is set to 256.

\subsubsection{Text-conditioned Global Sampler}
Intuitively, zero-shot text-driven synthesis can be achieved by directly optimizing the generative model using the guidance from CLIP~\cite{patashnik_styleclip_2021, liu_fusedream_2021}. However, such a strategy is not robust in generating large-scale scenes, \textit{e.g.,} the top two rows of \textbf{Fig. \ref{fig:text-ldr}}, as the model has no sense of the holistic semantics. 
Therefore, we need to condition the generation process on holistic semantics derived from input texts. Driven by this observation, we propose a text-conditioned global sampler that samples features from the global codebook, to align the text with scenes in a language-free training. 

Our key insight is to align the sampler with the text by unsupervised learning techniques, \textit{i.e.,} K-nearest neighbors and contrastive learning. 
The idea is to generate a text condition $C_\mathrm{txt}$ that aims to approximate the real semantic meaning of the target panorama $I$ without text-panorama paired training data. Denote $E_\mathrm{txt}$ and $E_\mathrm{img}$ as the text encoder and image encoder of CLIP~\cite{radford_learning_2021}, respectively. Inspired by \cite{zhou_lafite_2022}, we first perturb the image feature $E_\mathrm{img}(I)$ to generate a pseudo text feature:
\begin{equation}
    \hat{C}_\mathrm{txt} = (1-\alpha) E_\mathrm{img}(I) + \alpha \epsilon||E_\mathrm{img}(I)||_2/||\epsilon||_2,
\end{equation}
where $\alpha$ is a fixed hyper-parameter and $\epsilon \sim \mathcal{N}(0, \mathbf{I})$ is the Gaussian noise. Furthermore, we fetch the top-K nearest image embeddings $C_\mathrm{knn} = \{E_\mathrm{img}(I_k)\}^K_{k=1}$ of $\hat{C}_\mathrm{txt}$ as the additional condition. Consequently, the text condition $C_\mathrm{txt}$ is written as:
\begin{equation}
\label{eq:align-text}
C_\mathrm{txt} = \{C_\mathrm{knn} | \hat{C}_\mathrm{txt}\}.
\end{equation}

Specifically, we adapt the transformer~\cite{vaswani2017attention} into a text-conditioned sampler, which samples holistic features from the global codebook $\mathcal{Z}_g$ given the input text $T$.
We train the sampler conditioned on $C_{txt}$ in an autoregressive manner. The discrete representation $z_g$ is equivalent to a sequence $s \in \{0, ..., |\mathcal{Z}_g| - 1\}^{8\times 16}$ of indices from the holistic codebook $\mathcal{Z}_g$. Sampling global features is formulated as autoregressive next-index prediction conditioned on $C_\mathrm{txt}$. Given indices $s_{<i}$, the transformer-based sampler learns to predict the distribution of possible next indices, \textit{i.e.,} $p(s_i|s_{<i}, C_\mathrm{txt})$, to compute the likelihood of the full representation as $p(s)=\prod_ip(s_i|s_{<i}, C_\mathrm{txt})$. We maximize the likelihood to obtain a text-aligned sampler:
\begin{equation}
\label{eq:text-align-loss}
\mathcal{L}_\mathrm{txt} = \mathbb{E}_{I\sim p(I)}[-\log p(s)] + \mathcal{L}_\mathrm{con},
\end{equation}
where $\mathcal{L}_\mathrm{con}$ is the contrastive regularization:
\begin{equation}
\label{eq:contra}
    \mathcal{L}_\mathrm{con} = -\tau \sum^n_{i=1}\log \frac{\mathrm{exp}((E_\mathrm{img}(I_i) \cdot \hat{C}_\mathrm{txt})/\tau)}{\sum^n_{j=1}\mathrm{exp}((E_\mathrm{img}(I_j)\cdot \hat{C}_\mathrm{txt})/\tau)},
\end{equation}
where we maximize the cosine similarity between the real image feature and the pseudo text feature to reduce the gap of the joint text-image space of CLIP. $\tau \in (0, 1]$ is a hyper-parameter. 

Trained by Eqn. \ref{eq:text-align-loss}, the sampler is aligned with real text features, which enables zero-shot transferability by replacing $\hat{C}_\mathrm{txt}$ with $E_\mathrm{txt}(T)$ during inference (\textbf{Fig. \ref{fig:stage1}}). Intuitively, as the global codebook represents the full panorama at low resolution ($256\times 128$), our text-conditioned global sampler takes a glance at the entire scene and outputs the holistic condition $z_g$ to guide the following step on local patch synthesis.

\begin{figure}[t]
    \begin{center}
    \centerline{\includegraphics[width=1.01\columnwidth]{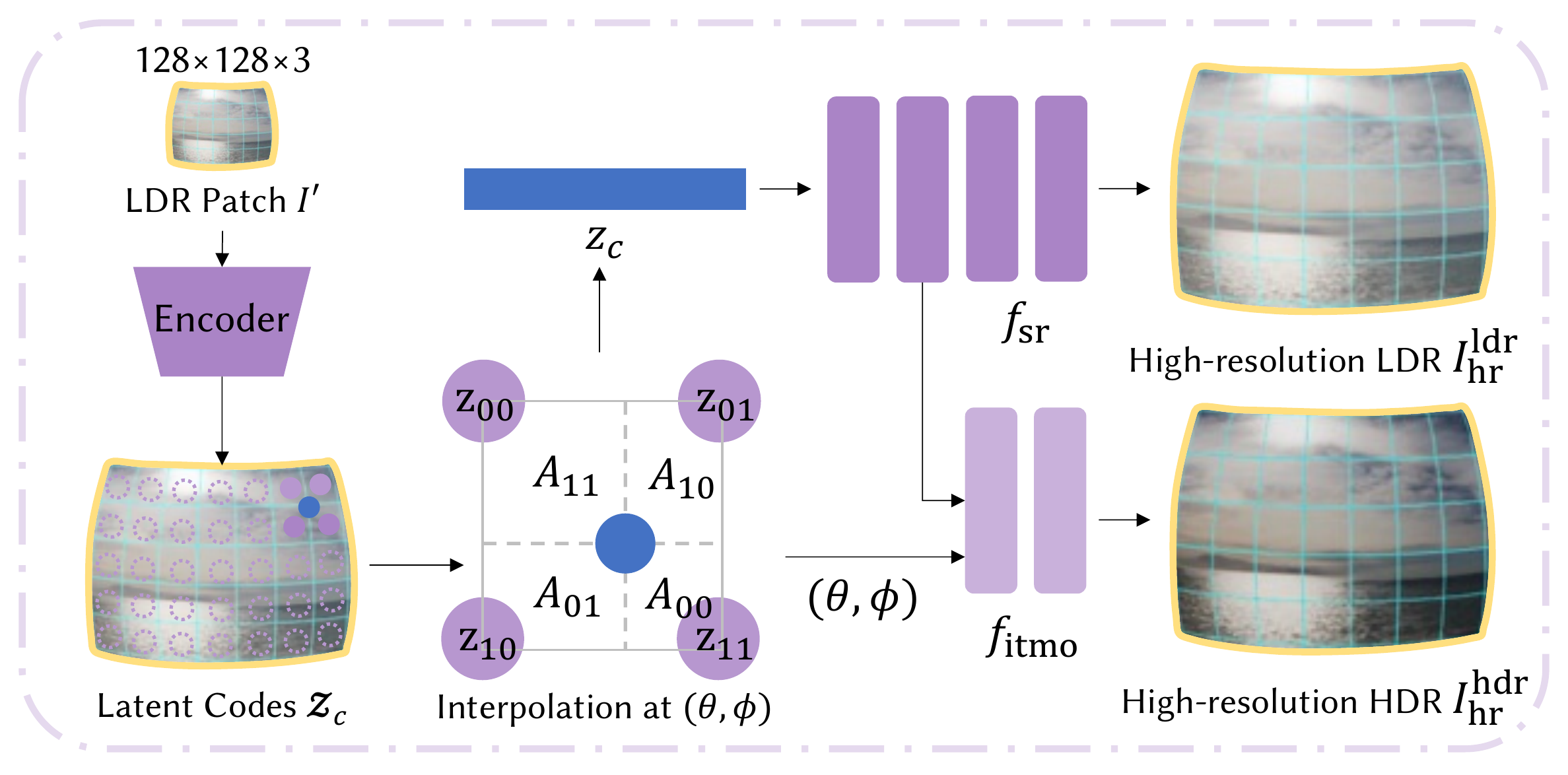}}
    \caption{\textbf{Overview of Stage II}. In Stage II, we upscale the result from Stage I both in spatial resolution and dynamic range using a continuous representation. For any position on the sphere $(\theta, \phi)$ in a continuous domain, we derive the latent codes $\mathcal{Z}_c$ and get the feature vector $z_c$ via area-based interpolation. Then, simple MLPs are adopted to query HDR values at $(\theta,\phi)$.
    }\label{fig:stage2}
    \end{center}
\end{figure}

\subsubsection{Structure-aware Local Sampler}
Given the holistic condition $z_g$, we focus on synthesizing local patches with sufficient fine-grained details. Thanks to the local codebook $\mathcal{Z}_l$, we are able to generate individual patches with high fidelity. Furthermore, the synthesizer needs to be aware of the intrinsic structure of panorama to properly composite patches according to the holistic condition, where the spherical positional encoding (Sec. \ref{sec:spe}) comes in.

In specific, we employ another transformer as a structure-aware sampler that samples features from the local codebook $\mathcal{Z}_l$. Given the holistic condition $z_g$, we want to synthesize the local patch $I_l \in \mathbb{R}^{256\times 256\times3}$ with spherical representation $S(\theta_l, \phi_l, 1) $, where $\theta_l \in \mathbb{R}^{256\times 256}, \phi_l \in \mathbb{R}^{256\times 256}$ are derived from Eqn. \ref{eq:spherical}. The discrete representation of $I_l$ is $z_l$ which is equivalent to a sequence $s \in \{0, ..., |\mathcal{Z}_l| - 1\}^{16\times 16}$ of indices from the local codebook $\mathcal{Z}_l$. 
SPE is directly prepended to the top of the sequence $s$ after the input tokens are mapped to embeddings. The corresponding spherical coordinates of $s$ is a vector with a dimension of $2\times 256 = 512$. And the SPE with Fourier features according to Eqn. \ref{eq:spe} is derived and stacked to form a tensor with a size of $18\times 512$, which is then concatenated with the embeddings in the transformer.
The structure-aware sampler learns to predict the next possible indices, \textit{i.e.,} $p(s_i|s_{<i}, z_g, \gamma(\theta_l), \gamma(\phi_l))$. We maximize the log-likelihood over the full representation:
\begin{equation}
    \mathcal{L}_\mathrm{local} = \mathbb{E}_{I_l\sim p(I_l)}[-\log p(s)],
\end{equation}
where $p(s) = \prod_i p(s_i|s_{<i}, z_g, \gamma(\theta_l), \gamma(\phi_l))$. 

During inference, we utilize this sampler in a sliding-window manner, as illustrated in \textbf{Fig. \ref{fig:stage1}}. For example, to generate a full panorama with the holistic condition $z_g$, we first generate its SPE as $\theta \in \mathbb{R}^{512\times 1024}$ and $\phi \in \mathbb{R}^{512\times 1024}$. Then, we patch-wisely slide the attention window of the sampler with size of $256\times 256$ to obtain the full representation $z_l \in \mathbb{R}^{32\times 64\times c_z}$. Finally, the full panorama in 1K resolution can be generated using the decoder of the local codebook, \textit{i.e.,} $I = G_l(z_l)$.

\subsection{Stage II: Super-Resolution Inverse Tone Mapping}
\label{sec:stage2}

In Stage II, \framework\ upscales the result from Stage I both in spatial resolution and dynamic range and outputs a high-quality HDR panorama in 4K+. On complementary of the discrete representation in Sec. \ref{sec:discrete}, we build a continuous representation that can be used to query HDR values at any spherical position on $S$:
\begin{equation}
\label{eq:continuous}
S(\theta, \phi, 1) = f_c(z_c, \theta, \phi),
\end{equation}
where $z_c$ is the feature derived from the LDR patch, and $S(\theta, \phi, 1)$ is introduced in Sec. \ref{sec:spe}. An overview of Stage II is shown in \textbf{Fig. \ref{fig:stage2}}.

\subsubsection{Structured Latent Codes as Continuous Representations}
The feature $z_c$ in Eqn. \ref{eq:continuous} is derived from a set of latent codes $\mathcal{Z}_c$ which represents the features distributed in the continuous domain of the sphere $S$. Given an LDR panorama $I$, we randomly crop an image patch $I' \in \mathbb{R}^{128\times 128\times 3}$ whose corresponding field of view on the sphere is $S'(\theta', \phi', 1)$. Then the structured latent codes $\mathcal{Z}_c$ are derived as a pixel-aligned feature map 
$\mathcal{Z}_c = E(I') \in \mathbb{R}^{128\times 128\times c_z'}$, where $c_z'$ is the dimension of each latent code. Then we compute the feature vector $z_c$ at any $(\theta, \phi)$ as an area-based interpolation:
\begin{equation}
    z_c = \sum_{i\in\{00,01,10,11\}} \frac{A_i}{A}z_i, \;\; z_i \in \mathcal{Z}_c,
\end{equation}
where $z_i$ is the nearest latent code at $(\theta, \phi)$, and $A_i$ is the area of the rectangle as illustrated in \textbf{Fig. \ref{fig:stage2}}.

\subsubsection{Super-Resolution Inversed Tone Mapping Operator}
With the help of the continuous feature $z_c$, we can leverage a composited function as $f_c = f_\mathrm{sr} \circ f_\mathrm{itmo}$, termed as Super-Resolution Inversed Tone Mapping Operator (SR-iTMO), to map value points in LDR patches to corresponding value points in high-resolution HDR patches. Specifically, the SR-iTMO consists of two tiny MLPs, $f_\mathrm{sr}$ and $f_\mathrm{itmo}$, that sequentially upscales the spatial resolution and dynamic range, as illustrated in \textbf{Fig. \ref{fig:stage2}}. $f_\mathrm{sr}$ is an MLP consisting of four layers, which upscales the LDR patch in resolution: 
\begin{equation}
\label{eq:sr}
    I_\mathrm{hr}^\mathrm{ldr}, c_\mathrm{hr} = f_\mathrm{sr}(z_c),
\end{equation}
where $c_\mathrm{hr}$ is the intermediate feature of the second layer of $f_\mathrm{hr}$. Then, $f_\mathrm{itmo}$, another MLP consists of two layers, upscales the dynamic range of the high-resolution patch:
\begin{equation}
\label{eq:itmo}
    I_\mathrm{hr}^\mathrm{hdr} = f_\mathrm{itmo}(c_\mathrm{hr}, \theta, \phi),
\end{equation}
where $\theta$ and $\phi$ are the spherical coordinates of the queried point.

\begin{figure}[t]
    \begin{center}
    \centerline{\includegraphics[width=1.01\columnwidth]{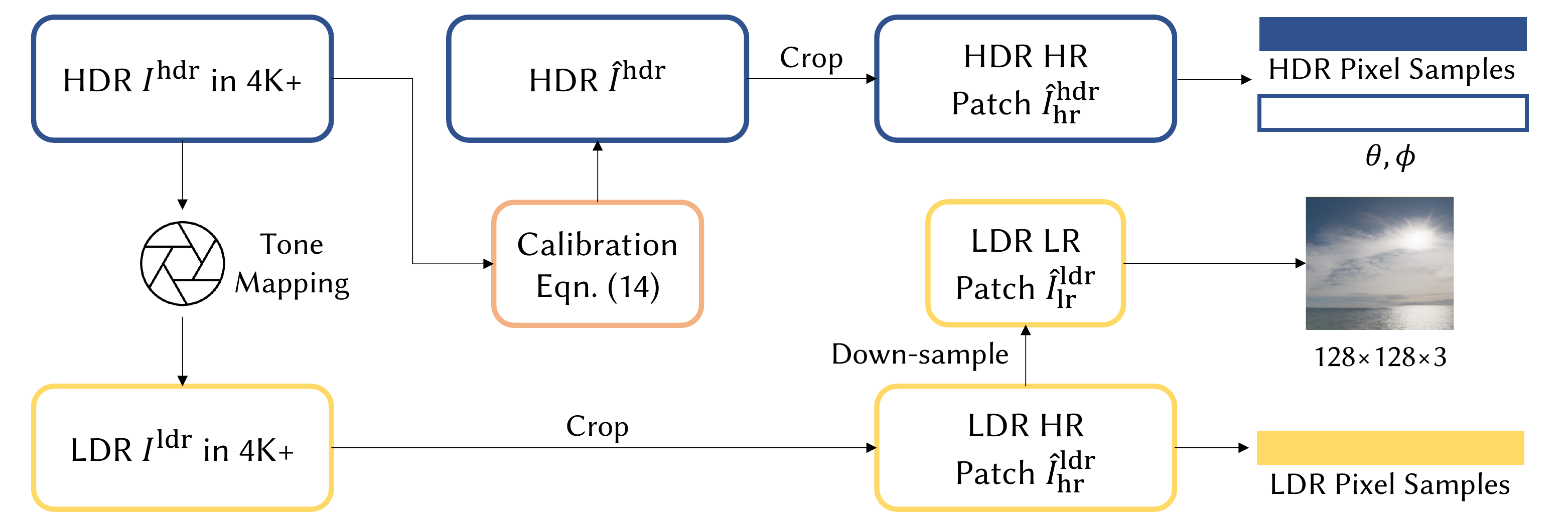}}
    \caption{\textbf{Data preparation and calibration of Stage II}. As \dataset\ contains high-resolution HDR panoramas ranging from 4K to 8K, we need to carefully generate training samples to train SR-iTMO (Sec. \ref{sec:train-itmo}), which can be decomposed into three steps which are tone mapping, calibration, and obtaining pixel samples, respectively.
    }\label{fig:data-pre}
    \end{center}
\end{figure}

\subsubsection{Data Preparation and Calibration}
To train our SR-iTMO module, whose goal is reconstructing HDR patches with higher resolution from LDR patches, we need to generate paired data, \textit{i.e.,} $\hat{I}_\mathrm{lr}^\mathrm{ldr} - \hat{I}_\mathrm{hr}^\mathrm{hdr}$ pairs, as training samples. We illustrate the whole process in \textbf{Fig. \ref{fig:data-pre}}.

First, we generate LDR panoramas $I^\mathrm{ldr} \in [0, 1]$ from HDR panoramas $I^\mathrm{hdr} \in [0, +\infty)$ via Reinhard tone mapping~\cite{reinhard}. Then, we calibrate raw HDR panoramas to remove scale ambiguity caused by different luminance scales across different scenes. Inspired by \cite{yu_luminance_2021}, we generate HDR ground truth based on the scale invariance of luminance. Given two HDR images $I^\mathrm{hdr}_1$ and $I^\mathrm{hdr}_2$, we define that they are luminance scale-invariant if and only if $I^\mathrm{hdr}_1 = \kappa I^\mathrm{hdr}_2, \kappa \in \mathbb{R}^{+}$. For each HDR panorama $I^\mathrm{hdr}$, we downscale it to its luminance scale-invariant version $\hat{I}^\mathrm{hdr}$ as the following:
\begin{equation}
\label{eq:calibration}
\hat{I}^\mathrm{hdr} = \frac{\sum(M \odot I^\mathrm{ldr})}{\sum(M \odot I^\mathrm{hdr})} \cdot I^\mathrm{hdr},
\end{equation}
where $\odot$ denotes the element-wise multiplication. And $M$ is a mask to determine the position of being calibrated based on $I^\mathrm{ldr} \in \mathbb{R}^{h\times w\times 3}$:
\begin{equation}
    M(i, j) = \left\{ 
\begin{aligned}
    &1, \;\; \sum_{k=1}^3 I^\mathrm{ldr}(i, j, k) < 3\sigma, \\
    &0, \;\;otherwise,
\end{aligned}
    \right.
\end{equation}
where $i = 0, 1, ..., h-1$ and $j = 0, 1, ..., w-1$. We set $\sigma=0.83$ in our experiments. 

Furthermore, we generate training samples at different resolutions. Here we take a single training pair as an example. Given a calibrated HDR panorama $\hat{I}^\mathrm{hdr}$, we randomly crop a patch $\hat{I}^\mathrm{hdr}_\mathrm{hr} \in \mathbb{R}^{128\beta \times 128\beta \times 3}$ as the high-resolution HDR, where $\beta \in [1,4]$ is a random scaling factor. The input LDR patch is generated by down-sampling $\hat{I}^\mathrm{ldr}_\mathrm{hr}$ to $\hat{I}^\mathrm{ldr}_\mathrm{lr} \in \mathbb{R}^{128\times 128\times 3}$. Note that, $\hat{I}^\mathrm{ldr}_\mathrm{hr} \in \mathbb{R}^{128\beta \times 128\beta \times 3}$ is cropped from the corresponding tone-mapped LDR panorama $I^\mathrm{ldr}$ at the same position as $\hat{I}^\mathrm{hdr}_\mathrm{hr}$. To match the size of pixel samples, we randomly choose $128\times 128$ points from $\hat{I}^\mathrm{hdr}_\mathrm{hr}$ and their corresponding spherical coordinates as high-resolution HDR ground-truth.

\subsubsection{Training}\label{sec:train-itmo}
We jointly train the encoder $E$ and the SR-iTMO $f_c$ in an end-to-end manner. Note that, with the help of our continuous representation, we are able to directly train the networks at any resolution. Our dataset contains HDR panoramas with resolutions ranging from 4K to 8K, and we do not scale them down to a fixed resolution during training. We first encourage the model to learn to increase the spatial resolution via minimizing:
\begin{equation}
    \mathcal{L}_\mathrm{sr} = \frac{1}{n}\sum||I^\mathrm{ldr}_\mathrm{hr} - \hat{I}^\mathrm{ldr}_\mathrm{hr}||_1,
\end{equation}
where $n$ is the number of pixel samples, and $I^\mathrm{ldr}_\mathrm{hr}$ is the network prediction from Eqn. \ref{eq:sr}. 

Besides, the model learns to expand the dynamic range by minimizing the following scale-invariant objective:
\begin{equation}
    \mathcal{L}_\mathrm{itmo} = \frac{1}{n}\sum D^2(I^\mathrm{hdr}_\mathrm{hr}, \hat{I}^\mathrm{hdr}_\mathrm{hr}) - [\frac{1}{n}\sum D(I^\mathrm{hdr}_\mathrm{hr}, \hat{I}^\mathrm{hdr}_\mathrm{hr})]^2,
\end{equation}
where $D(I_1, I_2) = \log(I_1) - \log(I_2)$ is a distance in logarithmic scale.

Thus, the overall learning objective of Stage II is a summation as $\mathcal{L}_\mathrm{full} = \mathcal{L}_\mathrm{sr} + \mathcal{L}_\mathrm{itmo}$. By this objective, the SR-iTMO learns to increase the resolution and dynamic range simultaneously.

\section{\dataset\ Dataset}

Currently, there is no publicly available dataset for \textbf{both} indoor and outdoor scenes. \cite{gardner2017learning} contributes an HDR dataset that contains only 2100 indoor scenes, and \cite{zhang_learning_2017} contributes around 200 outdoor HDR panoramas. To achieve a high-quality generative model, we newly collect thousands of panoramas from online resources and combine them with aforementioned datasets to form a new high-quality HDRI dataset, named \dataset, which contains 4392 HDR panoramas with resolutions ranging from $4096\times 2048$ (4K) to $8192\times 4096$ (8K).

Specifically, there are 1891 outdoor scenes and 2501 indoor scenes in \dataset. We further augment the data based on the intrinsic horizontal cyclicity of panoramas by clockwise rotating the panorama that wraps on the sphere ten times. Thus, the total amount of panoramas are 48,312. To facilitate text-driven scene synthesis, we also collect 1901 pieces of scene descriptions. Note that, in our experiments, these texts are only used as input during inference.

\begin{figure*}[p]
    \begin{center}
    \centerline{\includegraphics[width=2.0\columnwidth]{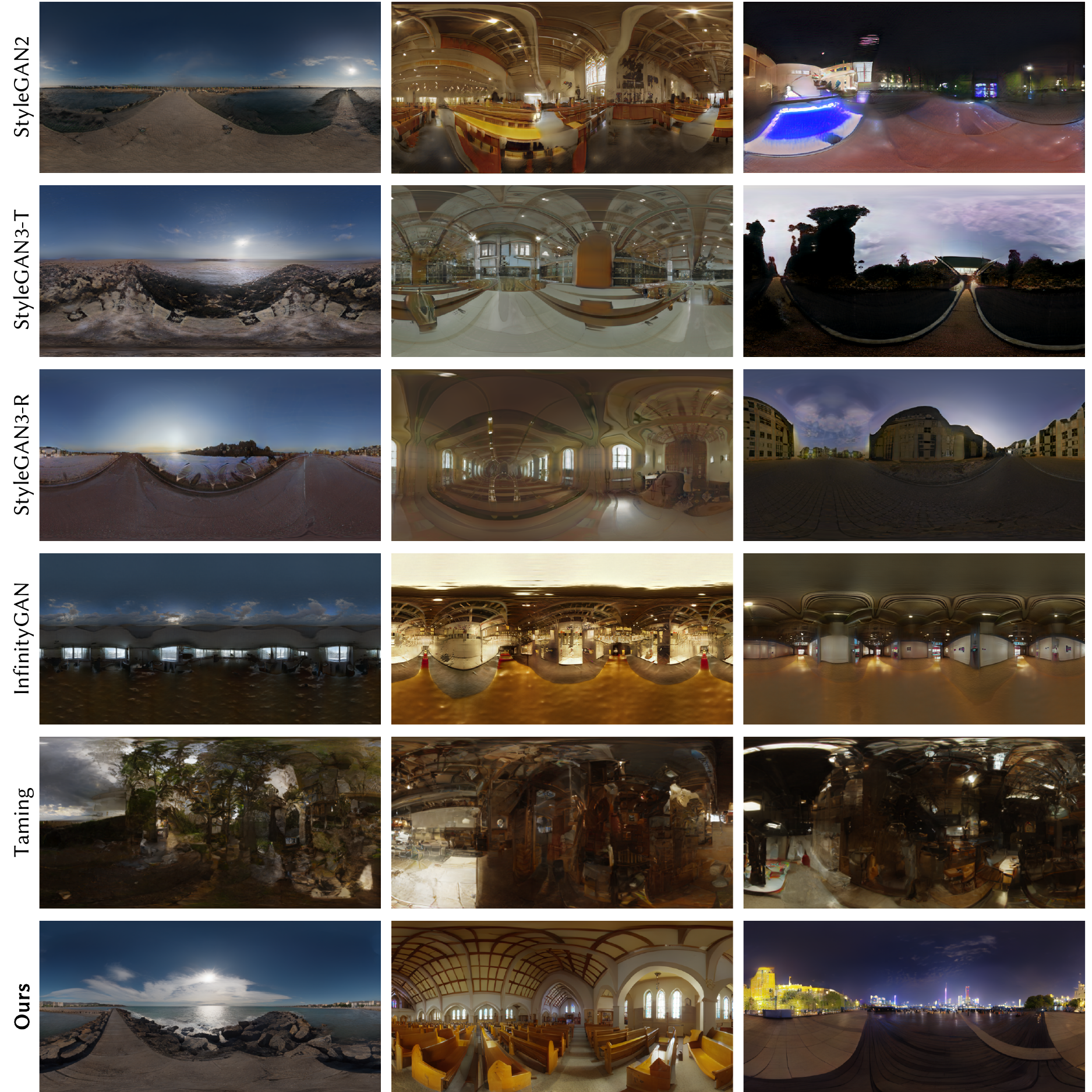}}
    \caption{\textbf{Visual Comparisons of LDR panoramas generation. Zoom in for details}. StyleGAN-based methods~\cite{karras_analyzing_2020, karras_alias-free_nodate} are directly trained on the full panorama with 1K resolutions. Their results seem to be realistic, yet contain local artifacts when zooming in. Plus, they fail to maintain the spherical structure of panoramic scenes. InfinityGAN~\cite{lin_infinitygan_2021} generates repeated patterns near the horizon while synthesizing textureless pixels near the pole. The coordinate-conditioned Taming Transformer~\cite{esser_taming_2021} fails to capture the diverse and structural intrinsic of panoramas. \framework\ succeeds in generating high-quality panoramas with sufficient details and diverse scenes.
    }\label{fig:ldr-gen}
    \end{center}
\end{figure*}

\section{Experiments}

\subsection{Implementation Details}
Our full model is trained stage by stage. We only utilize the tone-mapped LDR samples with 1K resolution in \dataset\ for the generation task in Stage I. And we split the dataset into a training set and a testing set for the inverse tone mapping task in Stage II. Specifically, the training set contains 45,056 samples, and the testing set contains 3,256 samples. We adopt Adam~\cite{kingma2014adam} optimizer for training.
\subsubsection{Stage I}
We first train the codebooks $\mathcal{Z}_\mathrm{g}$ and $\mathcal{Z}_\mathrm{l}$ for 60k iterations with a batch size of 12. Note that we replace all convolution layers of $E_\mathrm{g}$ and $G_\mathrm{g}$ with circular convolution layers in order to incorporate the cyclicity of the panorama. When training the text-conditioned global sampler, we freeze all parameters of CLIP~\cite{radford_learning_2021}. The hyper-parameter $\alpha$ is set to $0.25$, and the number of nearest neighbors is set as $K=5$. The text-conditioned global sampler is trained with a batch size of 16 for 100k iterations. Furthermore, we train the structure-aware local sampler with all other modules fixed. The batch size is set as 24.

\subsubsection{Stage II}
We leverage the backbone network in EDSR~\cite{lim2017enhanced} as the encoder $E$ to generate structured latent codes from LDR patches. $f_\mathrm{sr}$ is implemented as a four-layer MLP with 256 hidden units at each layer. $f_\mathrm{itmo}$ is another MLP that consists of two layers. During inference, we use the exponential function to map the output of $f_\mathrm{itmo}$ from the logarithmic scale to the original scale.

\begin{figure*}[t]
    \begin{center}
    \centerline{\includegraphics[width=2.0\columnwidth]{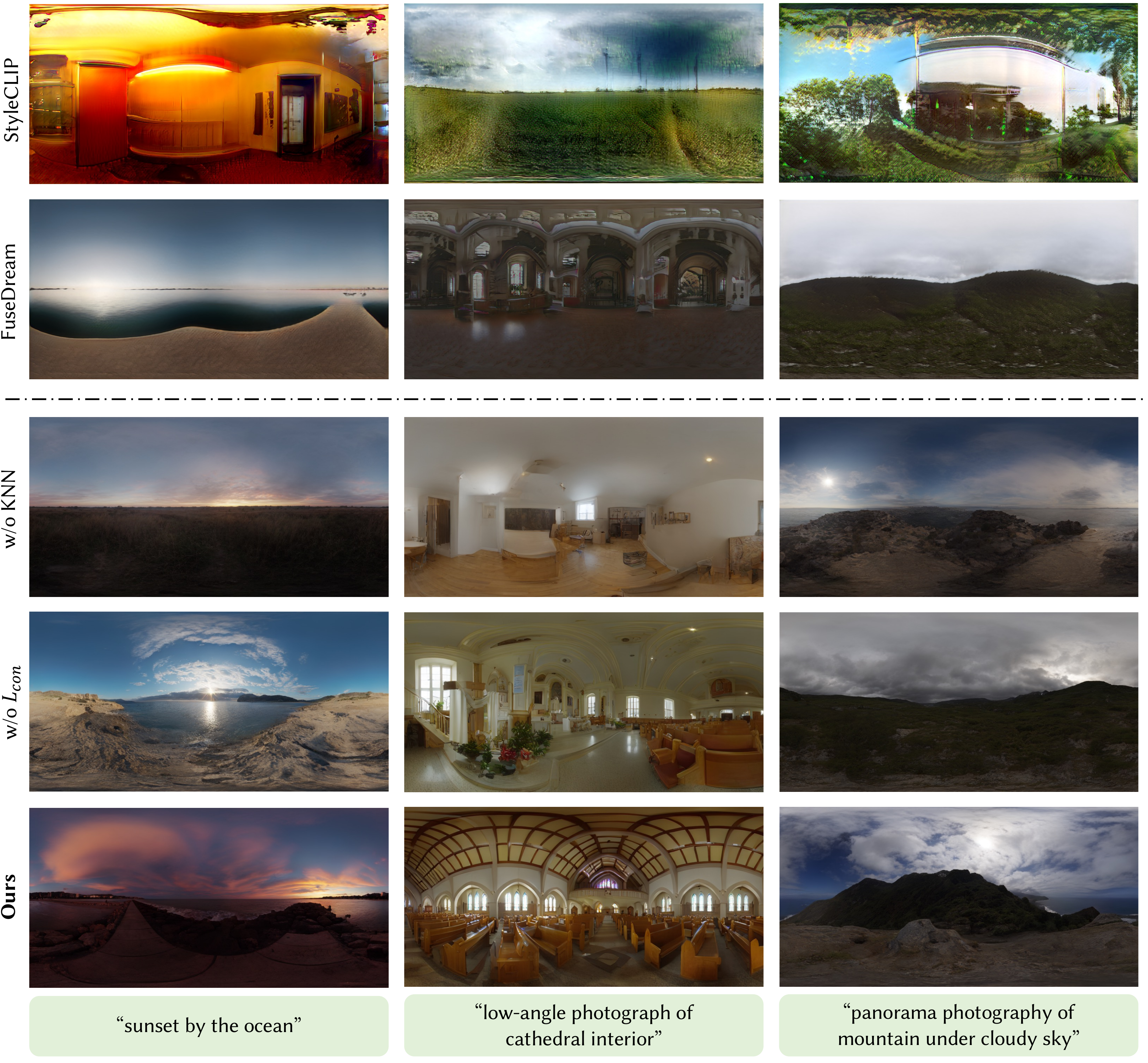}}
    \caption{\textbf{Visual Comparisons of zero-shot text-driven panorama synthesis. Zoom in for details}. (Above) \textbf{Comparison with baselines.} Compared with two optimization-based methods, the proposed method generates diverse textures in high-fidelity that perfectly match the input text, even for indoor scenes with high complexity. (Below) \textbf{Ablation studies.} The use of KNN (K-nearest neighbors) significantly improves the robustness of zero-shot text-driven synthesis. Plus, contrastive regularization helps the model better match the fine-grained details with specific texts, \textit{e.g.,} “sunset” with sunset clouds in the first column, and "body of water" in the third column.
    }\label{fig:text-ldr}
    \end{center}
\end{figure*}

\subsection{Comparison Methods and Evaluation Metrics}
For a fair comparison, we show our significant improvements over prior arts in panorama generation, zero-shot text-driven synthesis, and inverse tone mapping, respectively.

\subsubsection{Panorama Generation}\label{sec:ldr-exp}
We use Fréchet Inception Distance (\textbf{FID})~\cite{fid} and Inception Score (\textbf{IS})~\cite{salimans2016improved} as metrics to evaluate the quality of generated panoramas. 

Moreover, a user study is also conducted to evaluate the quality of panoramas, where each panorama is presented individually to the user for scoring. A total of 25 users are asked to: 1) score the perceptual quality (\textbf{PQ}), and 2) score the structural integrity (\textbf{SI}) of the panoramas. All scores are in the range of 5, and 5 indicates the best. We compare four state-of-the-art generators as follows:
\begin{itemize}
    \item \textbf{StyleGAN2}~\cite{karras_analyzing_2020} is a style-based generative model which synthesizes images from a latent distribution. We directly train it on the LDR panoramas with 1K resolution. Note that, we employ the adaptive augmentation technique~\cite{Karras2020ada} to stabilize training.
    \item \textbf{StyleGAN3}~\cite{karras_alias-free_nodate} is an alias-free version of StyleGAN2 whose internal representation is equivariant to translation and rotation. We consider two configurations (StyleGAN3-T and StyleGAN3-R) in our experiments.
    \item \textbf{InfinityGAN}~\cite{lin_infinitygan_2021} is a patch-based method that aims at arbitrary-sized image generation. We use the default setting to train their model on our LDR patches.
    \item \textbf{Taming Transformer}~\cite{esser_taming_2021} is a VQVAE-based generative model, succeeding in synthesizing high-resolution images with dense conditions, \textit{e.g.,} depth maps and segmentation maps. As there is no direct condition for HDR panoramas, we train a simple model that conditions on pixel coordinates.
\end{itemize}

\begin{table}[t]
\caption{\textbf{Quantitative results on LDR panorama generation.} The top three techniques are highlighted in red, orange, and yellow, respectively.
}\label{tab:gen}
\begin{tabular}{ccccc}
\toprule
     Methods& FID\ $\downarrow$& IS\ $\uparrow$& PQ\ $\uparrow$ & SI\ $\uparrow$ \\
     \midrule
     StyleGAN2&\cellcolor{oorange}14.62 &\cellcolor{yyellow}5.07$\pm$0.17&2.43&2.20\\
     StyleGAN3-T&16.61 &5.06$\pm$0.18&2.57&2.42 \\
     StyleGAN3-R&\cellcolor{yyellow}15.13 &5.01$\pm$0.19&2.19&1.87 \\
     InfinityGAN&104.62 &4.96$\pm$0.16&2.03&1.80\\
     Taming Transformer&112.38 &3.04$\pm$0.08&2.73&2.55\\
     \midrule
     w/o global&308.38&1.82$\pm$0.02&1.41&1.36\\
     w/o SP (Eqn. \ref{eq:spherical})&56.37&4.22$\pm$0.12&\cellcolor{oorange}3.45&\cellcolor{oorange}3.18\\
     w/o SPE (Eqn. \ref{eq:spe})&31.19&\cellcolor{oorange}6.32$\pm$0.27&\cellcolor{yyellow}3.23&\cellcolor{yyellow}3.09\\
     Ours&\cellcolor{rred}10.72 &\cellcolor{rred}6.65$\pm$0.22&\cellcolor{rred}4.26&\cellcolor{rred}4.25 \\
     \bottomrule
\end{tabular}
\end{table}

\subsubsection{Zero-shot Text-driven Synthesis}
We further evaluate the ability of our framework on zero-shot text-driven panorama synthesis, validating that simple extensions of previous methods cannot handle this challenging task. Same as Sec. \ref{sec:ldr-exp}, we use \textbf{FID} and \textbf{IS} to evaluate the quality of generation. Besides, another user study is performed for evaluation. The users are asked to: 1) score the perceptual quality (\textbf{PQ}), and 2) score the text consistency (\textbf{TC}) between the input text and the output panorama. All scores are in the range of 5, and 5 indicates the best. We compare two text-driven synthesis methods as follows:
\begin{itemize}
    \item \textbf{StyleCLIP}~\cite{patashnik_styleclip_2021} is a method that directly optimizes the latent space of StyleGAN using the guidance of CLIP. It iterates multiple times to maximize the cosine similarity of the generated image and the input text.
    \item \textbf{FuseDream}~\cite{liu_fusedream_2021} is an updated version of StyleCLIP, which applies augmentations on images to improve the robustness of aligning free-form texts to the generated image.
\end{itemize}

Note that, we use the StyleGAN2 model in Sec. \ref{sec:ldr-exp} as the underlying generator for these two methods.

\subsubsection{Inverse Tone Mapping}
We use Mean Absolute Error (\textbf{MAE}) and Root Mean Square Error (\textbf{RMSE}) as metrics to evaluate the performance of inverse tone mapping operators. 
Besides, to evaluate the quality of HDRIs in lighting 3D assets, we use them to render four shader balls with different materials, comparing the visual quality. We compare with three methods as follows:
\begin{itemize}
    \item \textbf{LANet}~\cite{yu_luminance_2021} is a luminance attentive network for HDR reconstruction from a single LDR image, which uses convolutional networks with a branch to output segmentation masks of the area with high luminance.
    \item \textbf{ExpandNet}~\cite{marnerides_expandnet_2019} is a fully convolutional model with three branches. It decomposes the image into local, global, and dilate features, which are fused at the bottom of the network.
    \item \textbf{HDR-CNN}~\cite{eilertsen_hdr_2017} is a convolutional autoencoder for HDR reconstruction. Its encoder converts an LDR input to a latent feature representation, and the decoder reconstructs this into an HDR image in the log domain.
\end{itemize}

\subsection{Evaluations on LDR Panorama Generation}
We first report the quantitative results in terms of LDR panoramas generation compared with representative methods, as shown in Table \ref{tab:gen}. Note that, three variants of StyleGAN are directly trained on the full panorama with 1K resolution. The rest two baselines are patch-based during training, and only synthesize the full panorama during evaluation. In a nutshell, \framework\ achieves the lowest FID and the highest IS among all methods, which demonstrates the high quality of our generation. Plus, the result of user study indicates that our method achieves the best perceptual quality to human users while also maintaining the structural coherence of panoramic scenes.

Furthermore, we qualitatively compare our method with baselines in \textbf{Fig. \ref{fig:ldr-gen}}. Although three variants of StyleGAN, which are trained on the full-framed panoramas, achieve relatively low FID metric, they cannot accurately catch the intrinsic structural property of panorama, leading to low fidelity. For example, the church interior synthesized by StyleGAN2 seems to be realistic at first glance, but the ceiling and bench are distorted and broken. InfinityGAN generates repeated patterns near the horizontal areas while synthesizing textureless pixels near the pole. The coordinate-conditioned Taming Transformer fails to synthesize the full panorama when trained on patches without any dense condition. The results are attributed to the lack of inductive bias of a plain transformer. In contrast, \framework\ learns to synthesize high-quality LDR panoramas thanks to careful designs on global semantic alignment and structure-aware local patch synthesis. 

\begin{figure*}[t]
    \begin{center}
    \centerline{\includegraphics[width=2.0\columnwidth]{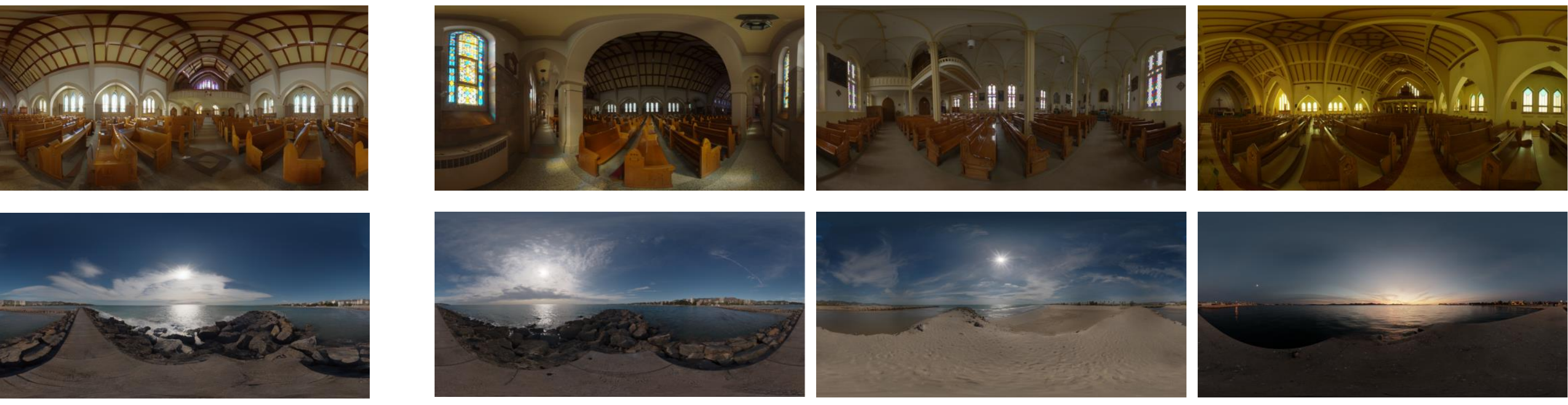}}
    \caption{\textbf{Nearest neighbors of two generated results, based on the cosine similarity in the image space of CLIP~\cite{radford_learning_2021}.} (Left) Panoramas synthesized by \framework. (Right) Nearest neighbors in \dataset\ measured by the cosine similarity between two images in the CLIP image space.
    }\label{fig:nn}
    \end{center}
\end{figure*}

In addition, we show nearest neighbors of our generated results based on the cosine similarity in the image space of CLIP~\cite{radford_learning_2021}, as shown in \textbf{Fig. \ref{fig:nn}}. The generated panoramas are not exactly identical to the training samples.

\begin{table}[t]
\caption{\textbf{Quantitative results on zero-shot text-driven synthesis.} The top three techniques are highlighted in red, orange, and yellow, respectively. 
}\label{tab:text}
\begin{tabular}{ccccc}
\toprule
     Methods& FID\ $\downarrow$&IS\ $\uparrow$& PQ\ $\uparrow$ & TC\ $\uparrow$  \\
     \midrule
     StyleCLIP &57.86&4.28$\pm$0.18&1.98&1.33\\
     FuseDream &\cellcolor{yyellow}38.16&\cellcolor{yyellow}5.89$\pm$0.28&\cellcolor{yyellow}2.41&\cellcolor{yyellow}1.57\\
     \midrule
     w/o KNN &52.85&4.77$\pm$0.23&2.14&1.23\\
     w/o $\mathcal{L}_\mathrm{con}$ &\cellcolor{oorange}34.70&\cellcolor{oorange}6.45$\pm$0.26&\cellcolor{oorange}3.81&\cellcolor{oorange}3.50\\
     Ours&\cellcolor{rred}32.01&\cellcolor{rred}6.46$\pm$0.37&\cellcolor{rred}4.38&\cellcolor{rred}4.56\\
     \bottomrule
\end{tabular}
\end{table}

\subsection{Evaluations on Zero-shot Text-driven Synthesis}
We further evaluate the capability of \framework\ in generating text-aligned content in high fidelity and zero-shot manner. The quantitative results compared with two CLIP+GAN methods are presented in Table \ref{tab:text}. Our careful design of the text-conditioned global sampler significantly improves the generation quality. 

Besides, the visual comparisons are shown in \textbf{Fig. \ref{fig:text-ldr}}. It is obvious that the simple combination of generative models and CLIP cannot catch the accurate semantics of the entire scene given the free-form text, which is validated by the failure of StyleCLIP~\cite{patashnik_styleclip_2021}. Indeed, the robust optimization process in FuseDream~\cite{liu_fusedream_2021} can somewhat improve the consistency between the content and the text. However, it still cannot convert all information from the text into a panoramic scene. For example, driven by the text "sunset by the ocean", FuseDream only synthesizes "ocean" but ignores the "sunset". In contrast, our model generates diverse textures that perfectly match the input text, even for indoor scenes with high complexity. Our hierarchical scheme enables the synthesizer both to align with the scene description globally and generate coherence complex details locally, \textit{e.g.,} cathedral interior with coherent structures of benches and ceilings. 

\subsection{Evaluations on Inverse Tone Mapping}
The quantitative comparisons on the testing set are presented in Table \ref{tab:itmo}. Note that, we set the upscaling factor $\beta=1$ of our method for fair comparisons. We validate the potential of realizing iTMO as MLPs to get rid of leveraging heavy fully convolutional architectures. Our iTMO achieves the lowest error on the testing set.

\begin{figure}[t]
\vspace{-0.1in}
    \begin{center}
    \centerline{\includegraphics[width=1.0\columnwidth]{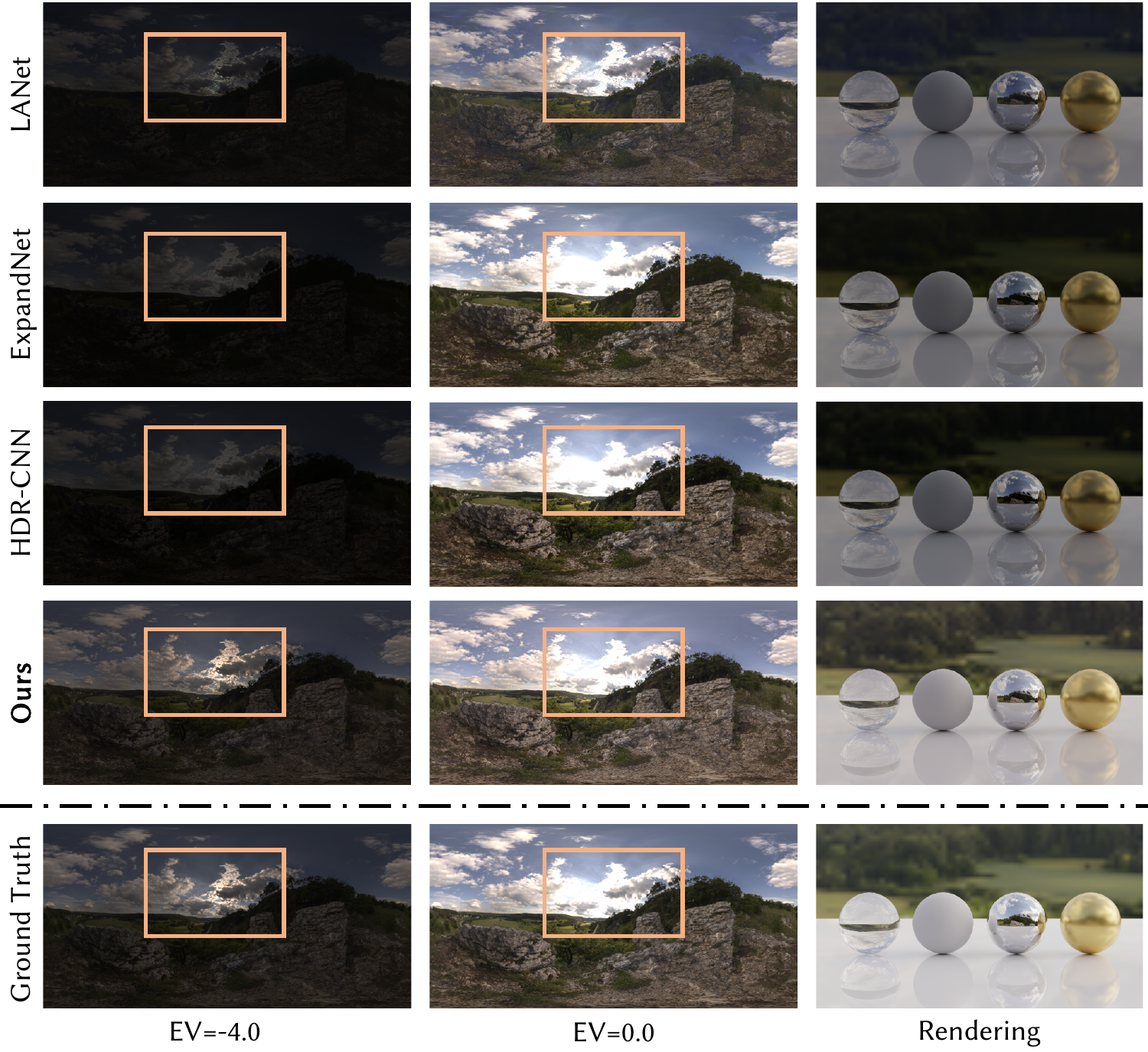}}
    \caption{\textbf{Visual Comparisons of iTMOs}. We qualitatively compare the predicted HDR panoramas at two exposure values (EV$=-4.0$, EV$=0.0$). In addition, we use the HDR panorama to render four shader balls to evaluate their capability for image-based lighting. The material of the four balls is glass, diffuse, glossy, and a mixture of diffuse and glossy, respectively.}\label{fig:itmo}
    \end{center}
\vspace{-0.1in}
\end{figure}

We make further visual comparisons to evaluate the quality of generated HDR panorama by different iTMOs, as shown in \textbf{Fig. \ref{fig:itmo}}. Compared with the ground-truth, only our method catches the high value of the radiance in the real world scene, validated as the bright region when the exposure value is $EV=-4.0$. The rendering results of the predicted HDR panoramas are also presented. We use four shader balls with different materials to evaluate the quality of panoramas for image-based lighting. Among all methods, ours is the closest one compared with ground truth. Different from iTMO for HDR displays, the one for HDR lighting needs to restore the scale of real-world luminance in order to produce photorealistic rendering, \textit{e.g.,} casting shadows and inducing specular highlights. 

Moreover, we demonstrate the versatility of our SR-iTMO in \textbf{Fig. \ref{fig:sritmo}}. With the benefits of the continuous representation $z_c$, the SR-iTMO can perform on arbitrary resolutions. Even the scaling factor $\beta$ (Sec. \ref{sec:train-itmo}) is only sampled within the interval of $[1, 4]$, we show the capability to scale up the LDR image eight times (x8), which can be simply achieved by increasing the resolution of the spherical coordinates $(\theta, \phi)$.

\begin{table}[t]
\caption{\textbf{Quantitative results on inverse tone mapping.} The top two techniques are highlighted in red, and orange, respectively.
}\label{tab:itmo}
\begin{tabular}{cccc}
\toprule
     Methods& MAE\ $\downarrow$& RMSE\ $\downarrow$\\
     \midrule
     LANet&0.3218&51.73\\
     ExpandNet&0.1894&26.68\\
     HDR-CNN&\cellcolor{oorange}0.1713&\cellcolor{oorange}26.40\\
     \midrule
     w/ single MLP&0.1798&31.36\\
     Ours&\cellcolor{rred}0.1442&\cellcolor{rred}26.38 \\
     \bottomrule
\end{tabular}
\end{table}

\subsection{Ablation Studies}

\subsubsection{Effectiveness of the global codebook}
We show the significant improvement brought by our hierarchical framework, especially the usage of the global codebook $\mathcal{Z}_g$. The term "w/o global", denotes a variant of our model where the global codebook is removed. As shown in Table \ref{tab:gen} and \textbf{Fig. \ref{fig:abl-ldr-gen}}, the quality generation drastically decreased. Since the local codebook and the structure-aware sampler are trained on image patches, there is no way that the model can directly synthesize full panoramas without knowing the global appearance. Plus, without the global codebook, our method loses the ability to align the text with the content.

\subsubsection{SPE is important for preserving structural coherence}
\begin{figure*}[t]
    \begin{center}
    \centerline{\includegraphics[width=2.0\columnwidth]{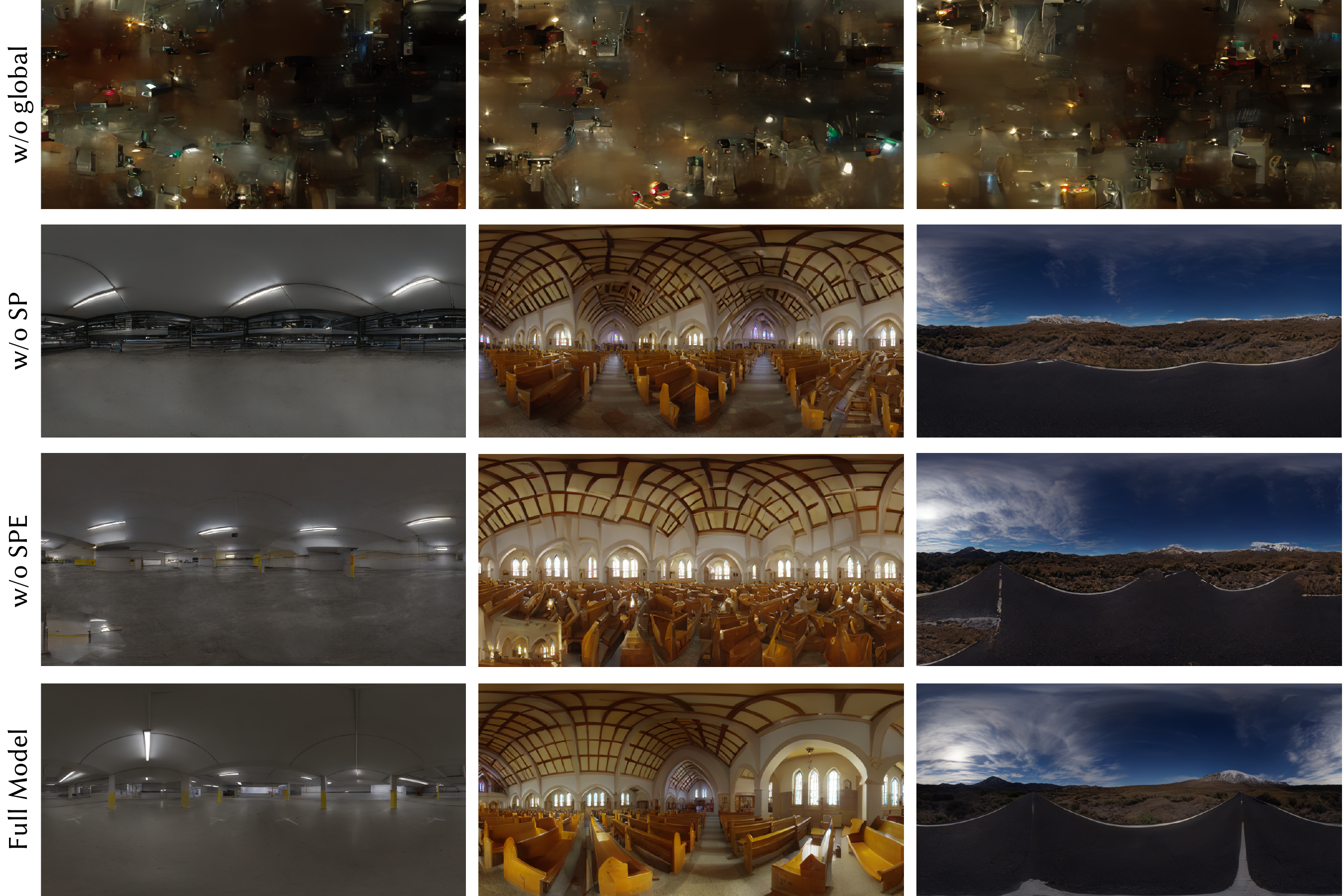}}
    \caption{\textbf{Qualitative ablation studies of LDR panoramas generation. Zoom in for details}. The global codebook and the corresponding sampling step is important for a global coherent generation. The spherical positional encoding (Eqn. \ref{eq:spherical} and Eqn. \ref{eq:spe}) is crucial for preserving structural coherence and unrepeated textures. Without such a representation, our model tends to generate repeated textures that expand horizontally.
    }\label{fig:abl-ldr-gen}
    \end{center}
\end{figure*}

To validate that our proposed SPE for panoramic images is crucial for preserving structural coherence, we implement two baselines. The first one, "w/o SP", directly removes representation in Eqn. \ref{eq:spherical} and Eqn. \ref{eq:spe}, and outpaints horizontally to synthesize full panoramas. The second one, "w/o SPE", only removes the Fourier feature of SPE in Eqn. \ref{eq:spe}. As shown in Table \ref{tab:gen}, our spherical representation of panoramas is important for generation in terms of overall quality and structural integrity. A visual explanation is presented in \textbf{Fig. \ref{fig:abl-ldr-gen}}. We could notice that, without SPE, our model tends to generate repeated textures. Plus, the encoding in Eqn. \ref{eq:spe} brings significant improvements in preserving the structural coherence, \textit{i.e.,} our full model generates continuous ceilings and lanes, which is the key to the fidelity of panoramic scenes.

\begin{figure}[t]
\vspace{-0.1in}
    \begin{center}
    \centerline{\includegraphics[width=1.0\columnwidth]{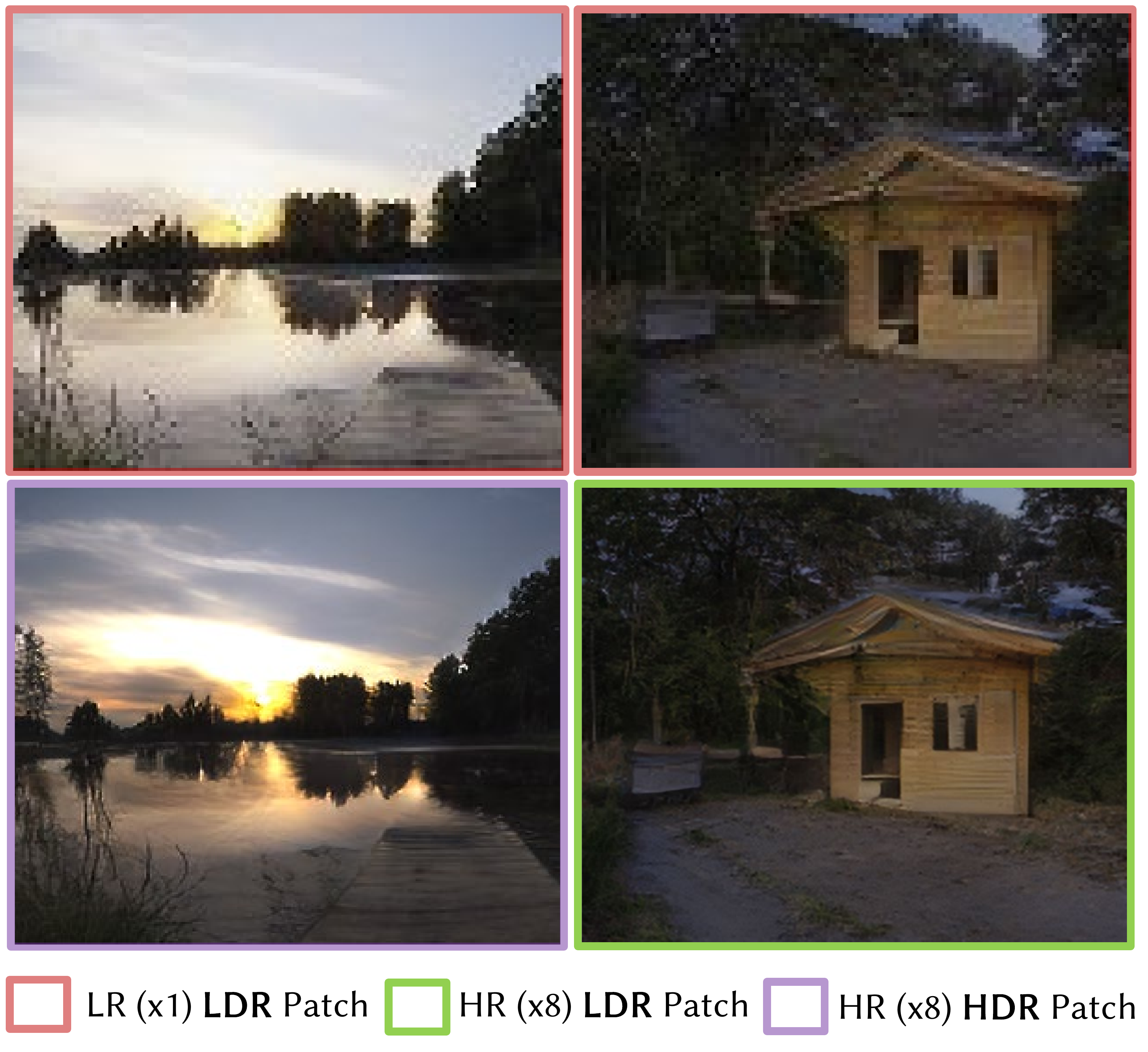}}
    \caption{\textbf{Illustration of our versatile SR-iTMO}. With the aid of our continuous representation, the SR-iTMO is capable of scaling up the LDR patch eight times (x8) to the HDR domain.
    }\label{fig:sritmo}
    \end{center}
\vspace{-0.1in}
\end{figure}

\subsubsection{Text-conditioned sampler improves zero-shot transferability}
To better align the free-form text with the generated scene via language-free training, we explore the unsupervised learning techniques, \textit{i.e.,} K-nearest neighbors (KNN in Eqn. \ref{eq:align-text}) and contrastive learning (Eqn. \ref{eq:contra}). We additionally train our model without these two techniques to ablate their impacts on zero-shot transferability. As shown in Table \ref{tab:text}, KNN significantly improves the generation quality and the consistency with the text, while the contrastive regularization brings a slight benefit. The qualitative results are presented in \textbf{Fig. \ref{fig:text-ldr}}. It is clear that KNN helps the model to eliminate generating artifacts when aligning with the text. Moreover, the contrastive regularization further improves \framework\ to locate and match the fine-grained details in text and images. 

\begin{figure}[t]
\vspace{-0.1in}
    \begin{center}
    \centerline{\includegraphics[width=1.0\columnwidth]{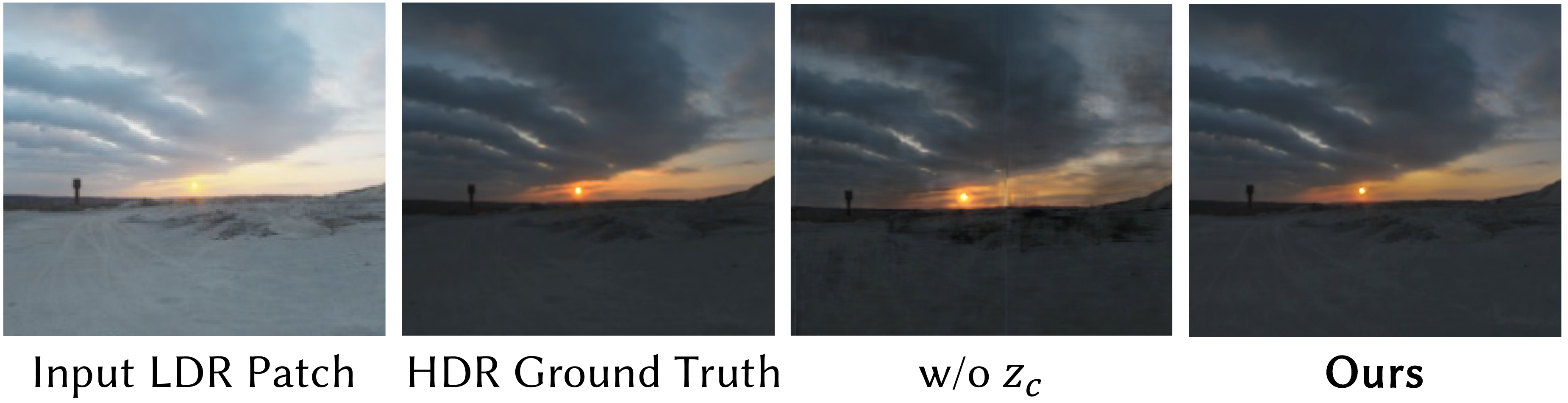}}
    \caption{\textbf{Ablation study of the continuous representation in Stage II}. Given an input LDR patch, the reconstructed HDR patch becomes blurry, and even contains checkerboard artifacts when our SR-iTMO is trained without $z_c$. We attribute this phenomenon to the lack of capability to learn a continuous representation that fits different scales.
    }\label{fig:abl-zc}
    \end{center}
\end{figure}

\begin{figure*}[tp]
    \begin{center}
    \centerline{\includegraphics[width=1.95\columnwidth]{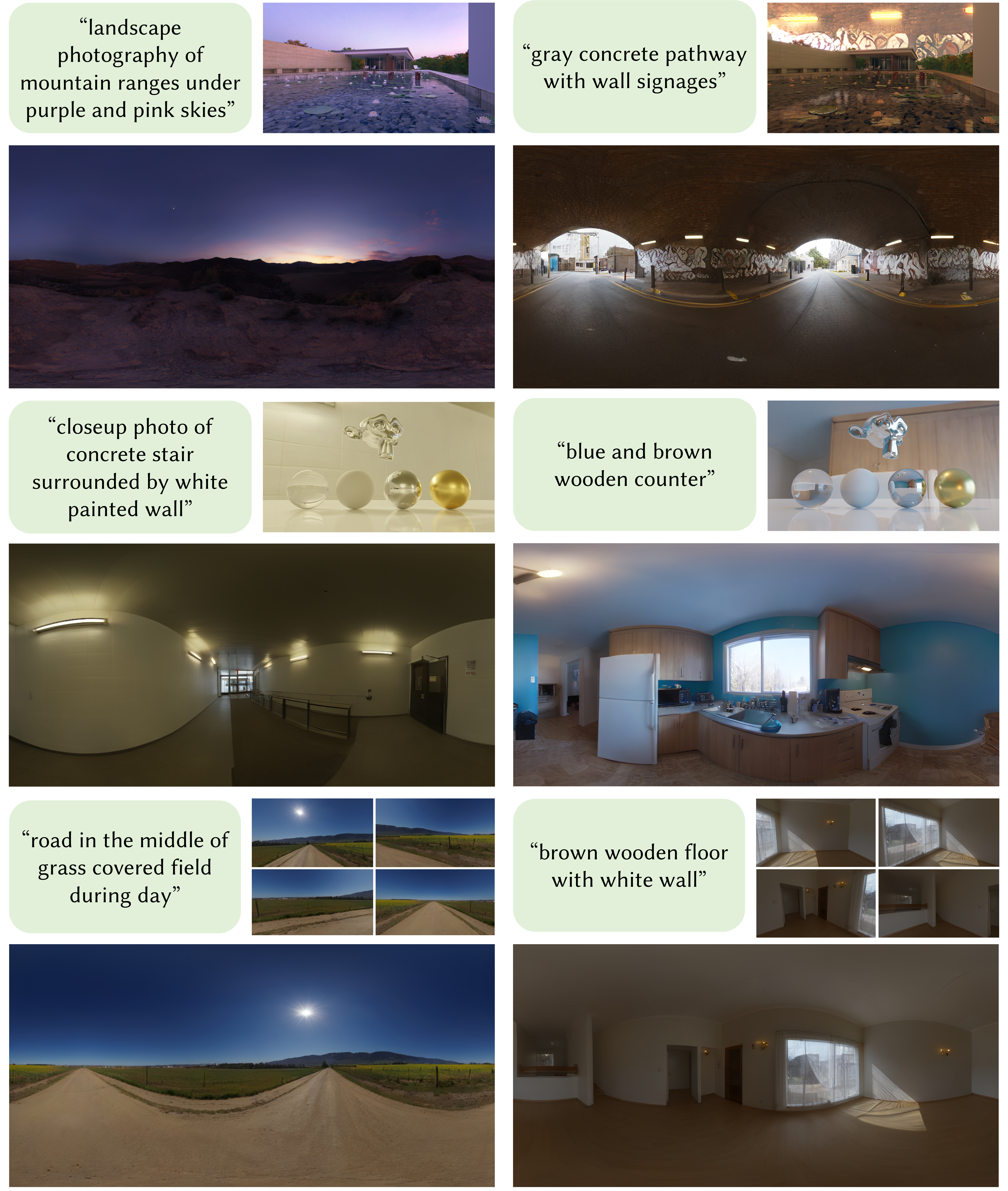}}
    \caption{\textbf{Applications of \framework}. \textbf{Top:} The generated HDR panoramas can be directly used in lighting 3D scenes to produce photorealistic reflections and illuminations. \textbf{Middle:} \framework\ can use text to light 3D assets with high fidelity. \textbf{Bottom:} The high quality of the generated panoramas guarantees the immersive experience for users to create a virtual world using solely natural language description.
    }\label{fig:application}
    \end{center}
\end{figure*}

\subsubsection{Benefits of the continuous representation $z_c$}
Besides the arbitrary scaling factor (\textbf{Fig. \ref{fig:sritmo}}), the continuous representation brings other benefits to our SR-iTMO. There is a naive way to realize our SR-iTMO, which directly learns the coordinate-based MLPs $f_\mathrm{sr}, f_\mathrm{itmo}$ without using continuous latent representation $z_c$. We show the quantitative comparison in \textbf{Fig. \ref{fig:abl-zc}}. Given the same input LDR patch, one can observe that the reconstructed HDR contains blurry and checkerboard artifacts when the model is trained without $z_c$. We attribute this phenomenon to the lack of capability to learn a continuous representation that fits different scales.

\begin{figure}[t]
\vspace{-0.1in}
    \begin{center}
    \centerline{\includegraphics[width=0.98\columnwidth]{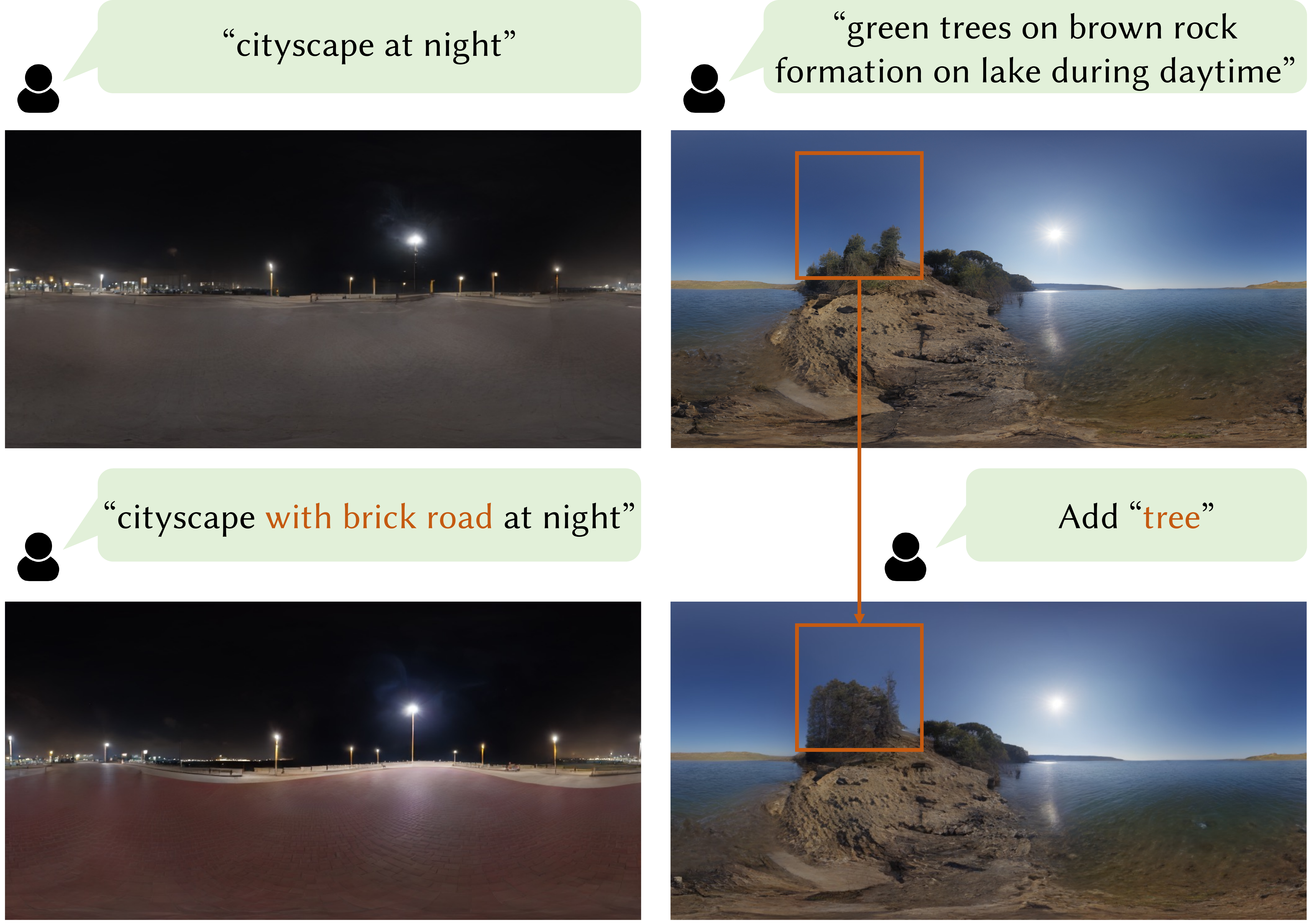}}
    \caption{\textbf{Scene editing by \framework}. With the hierarchical framework, we can further edit the generated scene by adding phrases or manipulating certain regions by extra descriptions. 
    }\label{fig:editing}
    \end{center}
\vspace{-0.1in}
\end{figure}

\subsubsection{Why Stage II uses two separated MLPs.}
Intuitively, treating the task of upscaling LDR images to super-resolved HDR images as a whole is always possible, which might motivate the use of a single MLP in Stage II. However, we turn to use separated MLPs, $f_\mathrm{sr}$ and $f_\mathrm{itmo}$, to model two different process independently. One is super-resolution while the other is inverse tone mapping. Separating two processes allows us to apply different losses to different branches to facilitate optimization. For example, the gradient from super-resolution loss $\mathcal{L}_{\mathrm{sr}}$ will not affect parameters of $f_{\mathrm{itmo}}$. We ablate this design choice by retraining a single MLP model with a comparable number of parameters as the original model, as reported in \textbf{Table \ref{tab:itmo}}. The use of two separated MLPs significantly improves the performance as the single MLP model lacks a disentangled modeling of super-resolution and inverse tone mapping. Besides, the physical process of inverse tone mapping has spatial bias. The radiance with high dynamic range, \textit{e.g.,} sun and lamp, often comes from the upper hemisphere of the spherical panorama. Thus, the spherical coordinates fed into $f_\mathrm{itmo}$ also serve as an inductive bias for inverse tone mapping but not super-resolution, which further motivates the use of two MLPs.

\subsection{Applications}

\subsubsection{Photo Realistic Rendering}
The generated panoramas can be directly used in modern graphic pipelines to produce photorealistic rendering on 3D assets. We present additional results in \textbf{Fig. \ref{fig:application}} for indoor and outdoor scenes, shown in the top two rows. Thanks to our careful designs in generating high-quality textures and inverse tone mapping, driven by only simple texts, \framework\ can render high-fidelity photos and provide realistic 360-degree reflections and illuminations.

\subsubsection{Immersive Virtual Reality}
Besides, our generation pipeline enables an amazing application that allows users to create their 3D scenes using only free-form texts, where users can view in any direction for an immersive VR experience. Two examples are presented in the bottom row of \textbf{Fig. \ref{fig:application}}, and more results will be shown in the supplementary videos.

\subsubsection{Scene Editing by Texts}
Thanks to our hierarchical generation framework, we can further manipulate and edit the generated scenes, as shown in \textbf{Fig. \ref{fig:editing}}. For example, by adding a phrase "with brick road" to "cityscape at night", we can change the appearance of the road in the panoramic scene. Plus, by changing the holistic condition of certain area, we add more "trees" to the generated rock.

\begin{figure}[t]
\vspace{-0.1in}
    \begin{center}
    \centerline{\includegraphics[width=0.98\columnwidth]{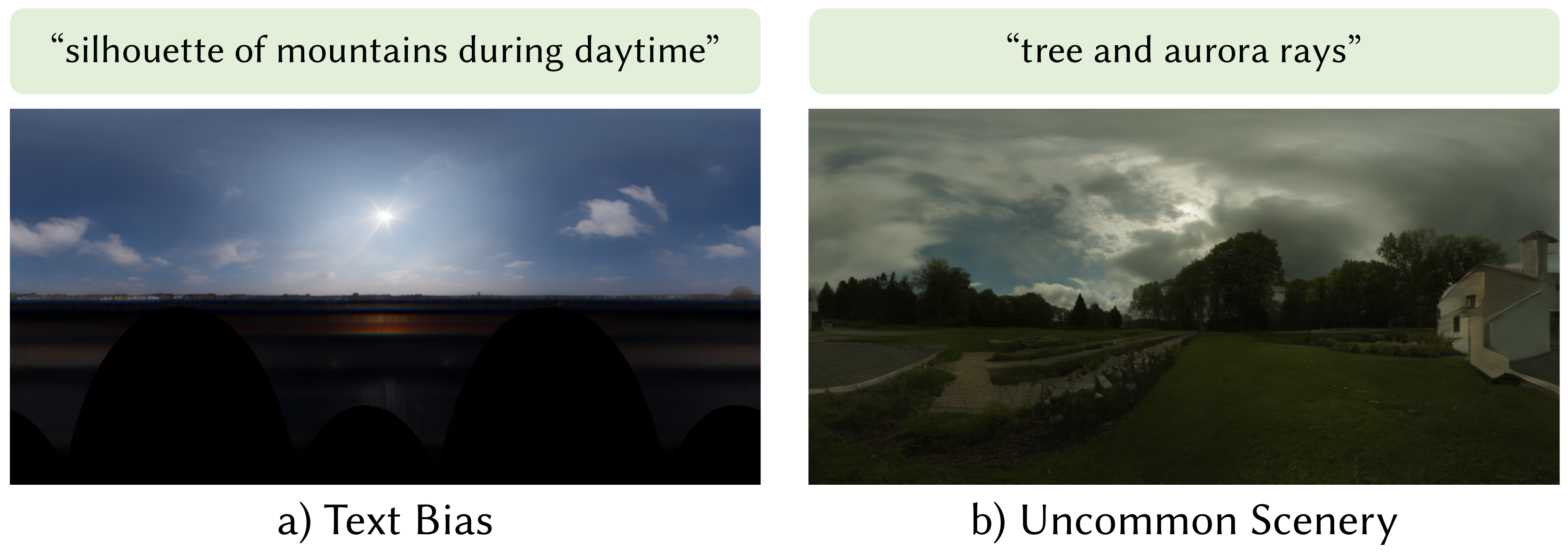}}
    \caption{\textbf{Failure cases}. a) The CLIP model is biased to the word "silhouette" which leads to pure dark regions that degrade the quality. b) Uncommon scenery, like "aurora", seems to be ignored during synthesis.
    }\label{fig:failure}
    \end{center}
\vspace{-0.1in}
\end{figure}

\section{Limitations}
In this section, we discuss two limitations of \framework. \textbf{1) Text Bias.} Since we do not use any paired data for text-driven synthesis, the performance relies on CLIP model~\cite{radford_learning_2021}. Interestingly, we observe that CLIP tends to be biased on some word-image pairs in our framework. For example, as shown in \textbf{Fig. \ref{fig:failure}(a)}, our generation pipeline is impacted by the bias of CLIP on the word "silhouette". If the word "silhouette" appears in the input description, our model will synthesize pure dark regions regardless of other words, which leads to a low fidelity. \textbf{2) Uncommon Scenery}. As presented in \textbf{Fig. \ref{fig:failure}(b)}, \framework\ cannot synthesize scene-level contents that are uncommon in the dataset. Therefore, when the sentence "tree and aurora rays" is given, our model only focuses on the term "tree". We attribute it to the use of KNN during training the text-align sampler. One possible solution would be further exploiting the joint image-text space of CLIP, which is still an open and challenging problem in zero-shot text-driven synthesis tasks.

\section{Conclusions}
This work proposes a zero-shot text-driven framework, \framework, to generate HDR panoramas with 4K+ resolution. Given a free-form text as the description of environment textures, we synthesize the corresponding HDR panorama with two steps: 1) text-driven LDR panorama generation and 2) super-resolution inverse tone mapping. Driven by the powerful CLIP model, we leverage a coarse-to-fine framework to generate LDR panoramas by free-form texts. Furthermore, by representing the panorama as a spherical field, we propose a versatile module to upscale the resolution and dynamic range of images jointly. Besides, we contribute a high-quality HDRI dataset, with rich and diverse scenes for the scene generation task. \framework\ shows the unprecedented ability of generating high-fidelity HDR panoramas in higher than 4K resolution and good alignment with the input text.

\begin{acks}
This work is supported by the National Research Foundation, Singapore under its AI Singapore Programme (AISG Award No: AISG2-PhD-2021-08-019), NTU NAP, MOE AcRF Tier 2 (T2EP20221-0012), and under the RIE2020 Industry Alignment Fund - Industry Collaboration Projects (IAF-ICP) Funding Initiative, as well as cash and in-kind contribution from the industry partner(s).
\end{acks}

\bibliographystyle{ACM-Reference-Format}
\bibliography{text2light}

\end{document}